\documentclass[10pt,twocolumn,letterpaper]{article}

\usepackage[pagenumbers]{iccv} 

\usepackage[dvipsnames]{xcolor}

\usepackage{stfloats}
\usepackage{utfsym}
\usepackage{booktabs}
\usepackage{multirow}
\usepackage{microtype}
\usepackage{tabularx}
\definecolor{iccvblue}{rgb}{0.21,0.49,0.74}
\usepackage[pagebackref,breaklinks,colorlinks,allcolors=iccvblue]{hyperref}
\usepackage[capitalize]{cleveref}
\usepackage[linesnumbered,ruled,vlined]{algorithm2e}
\newcolumntype{Q}[1]{>{\hsize=#1\hsize\centering\arraybackslash}X}
\usepackage[accsupp]{axessibility}

\def\paperID{8792} % *** Enter the Paper ID here
\def\confName{ICCV}
\def\confYear{2025}

\title{VoluMe -- Authentic 3D Video Calls from Live Gaussian Splat Prediction}

\author{Martin de La Gorce
\and 
Charlie Hewitt
\and 
Tibor Tak\'{a}cs
\and
Robert Gerdisch
\and
Zafiirah Hosenie
\and
Givi Meishvili
\and
Marek Kowalski
\and
Thomas J. Cashman
\and
Antonio Criminisi
\and\\
Microsoft, Cambridge, UK
}

\begin{document}

\twocolumn[{%
\maketitle
   \centering
\scriptsize
\captionsetup{type=figure}
\begin{tabularx}{0.495\linewidth}{@{}Q{0.166}@{}Q{0.833}@{}}
     Input frame & 3D reconstruction rendered from -40, -20, 0, +20, and +40 degrees
\end{tabularx}\hfill%
\begin{tabularx}{0.495\linewidth}{@{}Q{0.166}@{}Q{0.833}@{}}
     Input frame & 3D reconstruction rendered from -40, -20, 0, +20, and +40 degrees
\end{tabularx}\\
   \includegraphics[width=0.495\linewidth]{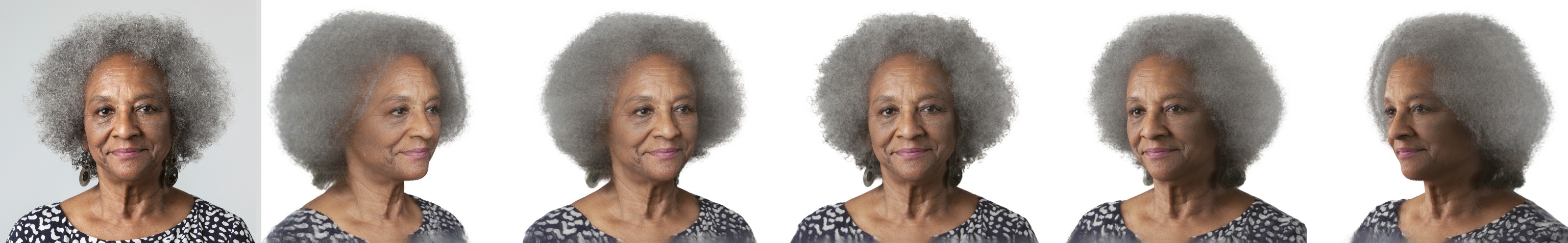}\hfill
   \includegraphics[width=0.495\linewidth]{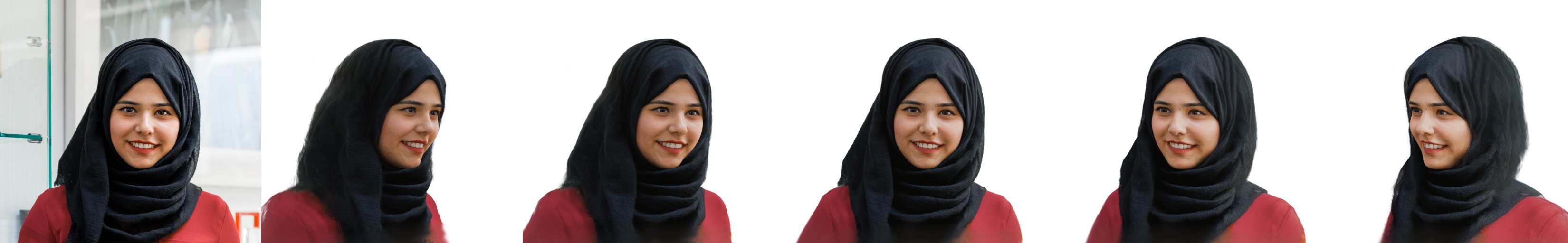}\\
   \includegraphics[width=0.495\linewidth]{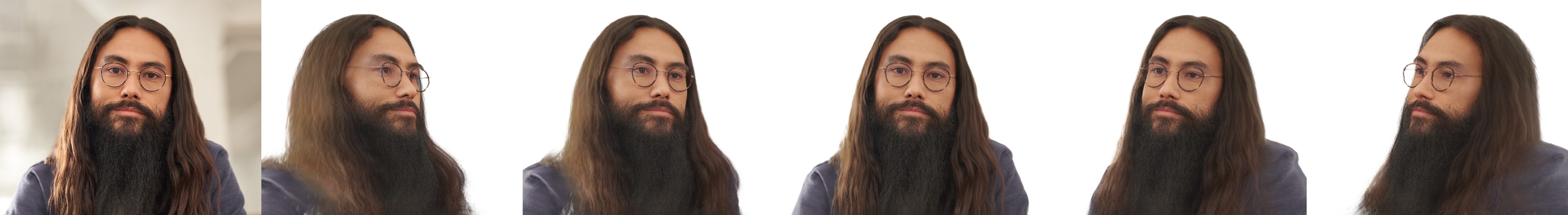}\hfill
   \includegraphics[width=0.495\linewidth]{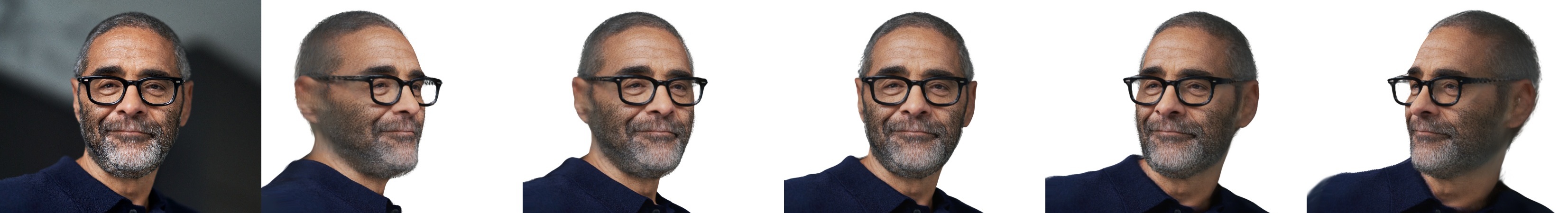}
   \\
   \includegraphics[width=0.495\linewidth]{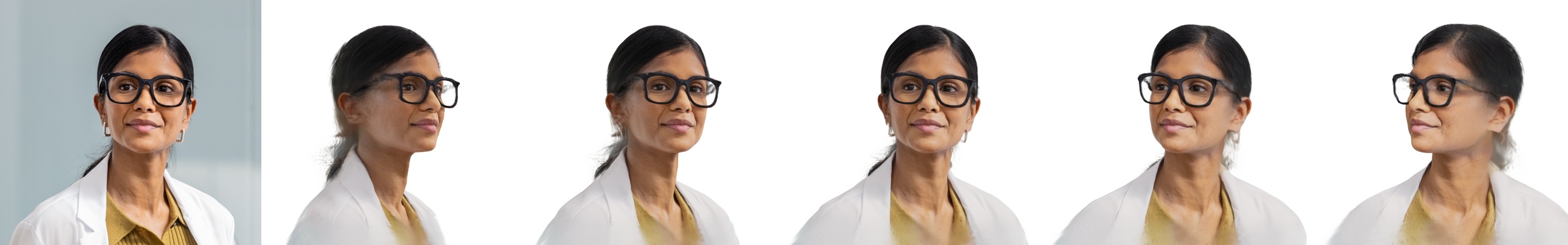}\hfill
   \includegraphics[width=0.495\linewidth]{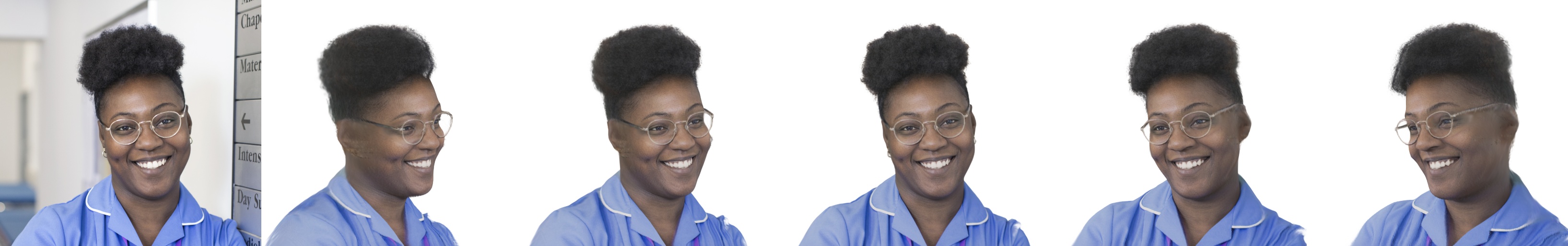}\\
   \captionof{figure}{
   Given a streaming monocular video of a person, our method reconstructs each frame as a realistic 3D Gaussian splat of the moment captured by that frame. Avatar systems build a model of the person's appearance in the past, and may not fully adapt to how a person looks in the present. Our method builds a realistic representation in real time and without enrolment. We use the term \emph{authenticity} for the property that the input frame is faithfully represented, shown in this figure by the match between the input frame and the reconstruction when rendered to the original camera's viewpoint. Our method faithfully represents eye glasses, headwear and diverse head and facial hair.
   }
   \vspace{10pt}
   \label{fig:teaser}
}]

\begin{abstract}

Virtual 3D meetings offer the potential to enhance copresence, increase engagement and thus improve effectiveness of remote meetings compared to standard 2D video calls. However, representing people in 3D meetings remains a challenge; existing solutions achieve high quality by using complex hardware, making use of fixed appearance via enrolment, or by inverting a pre-trained generative model. These approaches lead to constraints that are unwelcome and ill-fitting for videoconferencing applications.

We present the first method to predict 3D Gaussian reconstructions in real time from a single 2D webcam feed, where the 3D representation is not only live and realistic, but also \emph{authentic} to the input video. By conditioning the 3D representation on each video frame independently, our reconstruction faithfully recreates the input video from the captured viewpoint (a property we call \emph{authenticity}), while generalizing realistically to novel viewpoints. Additionally, we introduce a stability loss to obtain reconstructions that are temporally stable on video sequences.

We show that our method delivers state-of-the-art accuracy in visual quality and stability metrics compared to existing methods, and demonstrate our approach in live one-to-one 3D meetings using only a standard 2D camera and display. This demonstrates that our approach can allow anyone to communicate volumetrically, via a method for 3D videoconferencing that is not only highly accessible, but also realistic and authentic.

\end{abstract}
\section{Introduction}

Online video meetings have transformed the ability for people to communicate and collaborate while physically remote, but standard 2D video meetings create challenges with fatigue~\cite{raake_vc-fatigue-tech-factors_2022}, make it difficult to infer where attention is directed~\cite{schuessler-gazing-heads-2024},  and have limited effectiveness~\cite{olson-distance-matters-2000}.
Recent evidence suggests that virtual 3D meetings offer the potential to enhance copresence, increase engagement, and thus improve communication effectiveness~\cite{tang_perspectives_2022,schuessler-gazing-heads-2024,starline}.
However, representing people in 3D meetings, with accuracy and realism in both appearance and movement, and on consumer devices, is an unsolved challenge.

Existing solutions achieve high quality either with complicated hardware, such as multiple input cameras, dedicated sensors and powerful processors, e.g., Project Starline~\cite{starline}.
Alternatively, photoreal avatars~\cite{saito-gaussian-codec-avatars-2024} use a latent representation to animate a pre-enrolled model of the user, limiting the appearance to what was seen at enrolment time, and thus intentionally failing to adapt to the current moment.
Finally, for approaches such as Live 3D Portrait~\cite{trevithick2023} and TriPlaneNet~\cite{bhattarai2024triplanenet}, for which the 3D reconstruction is learned or predicted from the EG3D generative model~\cite{chan2022efficient}, the 3D reconstructions are limited by the textures and geometry learned by that model (see \cref{fig:qual_liv3d_fail}).
In each of these approaches, the constraints on the 3D representations are unwelcome for videoconferencing, where users need the rich, authentic and unconstrained representation that is familiar from 2D video, without the need for specialized hardware.
In order to 
meet this challenge, we need a 3D representation that is
\begin{enumerate}
    \item \textbf{Authentic:} generates images that match the input video foreground (with every detail of, for example, clothes, hair, and glasses from the present moment), when rendered from the original camera's viewpoint;
    \item \textbf{Realistic:} generates plausible and realistic reconstructions when rendered in new views, while supporting the full diversity of human communication (e.g.\ including diverse hair, headwear and accessories); 
    \item \textbf{Live:} runs in real time on consumer devices;% (we assume an NVIDIA GTX 4090);
    \item \textbf{Stable:} generates predictions that are stable with respect to time and as the viewpoint changes, to avoid flickering. 

\end{enumerate}
.
In this paper, we demonstrate the first method to satisfy all four properties.
Examples of our results are shown in \cref{fig:teaser}.
Note that we can do so \emph{without} the need for a full 360\textdegree{} reconstruction, while addressing an important class of 3D video meetings (the capture volume for Project Starline~\cite{starline}, for example, necessitates novel views only up to 55\textdegree{} from frontal).
In this paper, our contributions are:
\begin{itemize}
    \item We demonstrate 3D Gaussian reconstructions from 2D video that are live, authentic and realistic for a range of novel views that are up to 40\textdegree{} offset from the input.
    \item We show that these reconstructions are competitive with the state of the art for realism and quality, and excel at reconstructing the input view, even for a wide range of unconstrained inputs typical of videoconferencing scenarios (such as diverse hair, lighting, eyewear, clothing and accessories).
    We demonstrate state-of-the-art results on PSNR and LPIPS metrics on the Ava256 and Cafca datasets, and better results on jitter metrics on the Ava256 dataset.
    \item We demonstrate a real-time system that uses these 3D reconstructions to enable 3D videoconferencing at 30FPS with simulated motion parallax, using only a standard RGB camera and screen.
\end{itemize}

\subsection{Related work}

\noindent\textbf{Photorealistic avatars.}
Avatar-based techniques rely on an enrolment process, where an appearance and geometry model is built for a given user.
Enrolment may involve capturing multiple views of the user from a phone~\cite{cao-authentic-volumetric-2022,kirschstein2025avat3r} or multi-camera rig~\cite{saito-gaussian-codec-avatars-2024}; this process may be lengthy, as the user needs to show different head positions, facial expressions etc., and may be inaccessible or expensive in the case of multi-camera setups.
Recent work on photorealistic avatars relies on radiance field representations, first by adapting NeRF~\cite{Mildenhall20eccv_nerf} to represent a head model using deformation fields~\cite{Zielonka2022InstantVH} or deformable volumes~\cite{Lombardi21,garbin-voltemorph-2024,buehler2024cafca}.
With Gaussian splatting, \citet{kerbl3Dgaussians} demonstrate that radiance fields can be rendered more efficiently via rasterization, and inspire a large range of work that adapts Gaussian splatting to digital humans~\cite{qian2024gaussianavatars,xiang2024flashavatar,chen2023monogaussianavatar,xu2023gaussianheadavatar,giebenhain2023nphm,xu2024gphm,saunders2024gasp}.

In this work, we draw inspiration from the \emph{realism} demonstrated by Gaussian avatars, but apply this representation to a setting where \emph{authenticity} is important; we cannot rely on a fixed enrolled appearance from the past.
Instead we want to ensure that our 3D reconstruction adapts instantly to any appearance changes, as is true for all 2D video calls today.

\noindent\textbf{Single view reconstruction.} %
A second class of related work predicts a 3D reconstruction from a single 2D image or video frame, without prior enrolment.
These methods can offer high \emph{authenticity} and \emph{realism}% of both appearance and movement
, but the challenge of reconstructing 3D from a single 2D input makes it harder for these methods to be \emph{live} and \emph{stable}.

Live 3D Portrait~\cite{trevithick2023} and TriPlaneNet~\cite{bhattarai2024triplanenet} both make use of the EG3D generative model~\cite{chan2022efficient} and predict an implicit `triplane' radiance field representation. % from that model.
Live 3D Portrait runs in real time and achieves impressive results, satisfying many of our requirements, but by distilling knowledge from a generative 3D model, the resulting reconstructions are constrained by said model (see \cref{fig:qual_liv3d_fail}), and TriPlaneNet predicts a latent code for EG3D, making this a hard constraint.

The most promising explicit representation methods are those that predict Gaussian splat~\cite{kerbl3Dgaussians} outputs, as they can capture thin structures, specularities and transparencies.
\citet{xu2024agg} and \citet{Zou2023TriplaneMeetGS} combine the benefits of (implicit) triplane and (explicit) Gaussian splat representations, but both approaches are based on transformer architectures that do not scale well to \emph{live} high-resolution Gaussian splat prediction. 
\citet{lyu2024faceliftsingleimage3d} train a diffusion model to predict multiple views from a single image, which are then used to predict 3D Gaussians using GS-LRM~\cite{gslrm2024}; this grants high \emph{realism}, but the multi-view diffusion prediction makes it harder to produce \emph{stable} outputs for video sequences.

\citet{szymanowicz24splatter} circumvent these problems using a 2D convolutional U-Net architecture~\cite{song2021denoising} that directly predicts a Gaussian splat representation, or `splatter image', in a method that is capable of \emph{live} reconstruction but is not applied to human reconstruction, nor to video sequences.

In this paper, we build upon the Splatter Image~\cite{szymanowicz24splatter} method and demonstrate that direct Gaussian splat prediction can satisfy the requirements for \emph{realistic}, \emph{authentic}, \emph{live} and \emph{stable} 3D video calls.
In contrast to \citet{szymanowicz24splatter}, we use a planar homography to reduce perspective distortion when a person appears near the border of the input image,
we incorporate a scaling step during model training to deal with varying distances between the camera and the subject,
we output two Gaussians per pixel to increase the quality of the reconstruction,
and we add a stability loss to reduce temporal flicker in reconstructions of video sequences.
A concurrent work~\cite{szymanowicz2024flash3d} also uses two Gaussians per pixel, but for scene reconstruction rather than human reconstruction.

\noindent\textbf{3D reconstruction with synthetic data.} 
One interesting aspect of our work is the choice to train on purely synthetic data.
This enables multi-view data with perfect camera parameters that emulates in-the-wild scenarios; all important factors for our method.
However, recent work reconstructing 3D humans using methods trained on synthetic data have encountered issues generalizing to real images at test time.
For example, Rodin~\cite{wang2023rodin} trains a generative model to create 3D avatar representations.
Due to the synthetic training data their 3D results look noticeably synthetic, matching the domain of the training data.
Similar issues are visible for Live 3D Portrait~\cite{trevithick2023} and TriPlaneNet~\cite{bhattarai2024triplanenet} (e.g., eyes following the camera), although the EG3D-based synthetic data has a smaller domain gap to real images.
Other recent methods aim to explicitly handle the domain gap by training a generative prior model on synthetic data and fine-tuning on real enrolment data \cite{buehler2024cafca,saunders2024gasp}.
This does not suit our application, as the results are neither \emph{live} nor \emph{authentic}.
Instead we choose to follow Splatter Image~\cite{szymanowicz24splatter}, in which the output is much more directly tied to the input with a %more conventional
feed-forward network.
This allows our network to generalize well to real images, despite being trained only on synthetic data.
The skip connections and direct-sampling elements of the architecture are critical in allowing information to be forwarded directly from the input image to the output 3D representation, while the bottleneck can learn a strong prior over 3D geometry based on the synthetic training data.

\section{Training the 3D Reconstruction Model}
\label{sec:training}

\subsection{System Design}
Our method takes a single image as input, performs face detection on that image, extracts a region around the face and runs our model on this region.
The output of the model is a set of 3D Gaussians that represent the 3D shape and appearance of the face in the image.
The model architecture is based on Splatter Image~\cite{szymanowicz24splatter}.
We also use a U-Net architecture~\cite{ronneberger2015unet} to predict the Gaussians, though we choose a smaller backbone than SongUNet~\cite{song2021denoising} to reduce execution time.
We further reduce the number of convolution layers to only two per resolution and increase the number of resolutions to five to maintain a large receptive field, also omitting the self-attention layers.
To retain high-quality results we add four new ingredients in our method, described below.

\newcommand{\splatterfigheight}{1.95cm}

\begin{figure}
    \centering%
    \setlength{\tabcolsep}{1pt}
    \scriptsize
\begin{tabular}{ccccccc}
\includegraphics[height=\splatterfigheight]{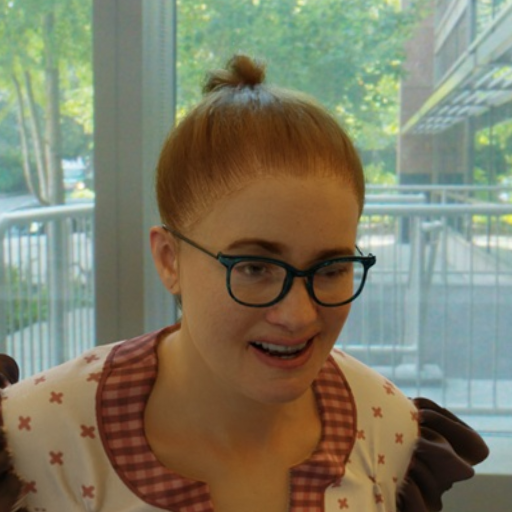}&
\includegraphics[height=\splatterfigheight]{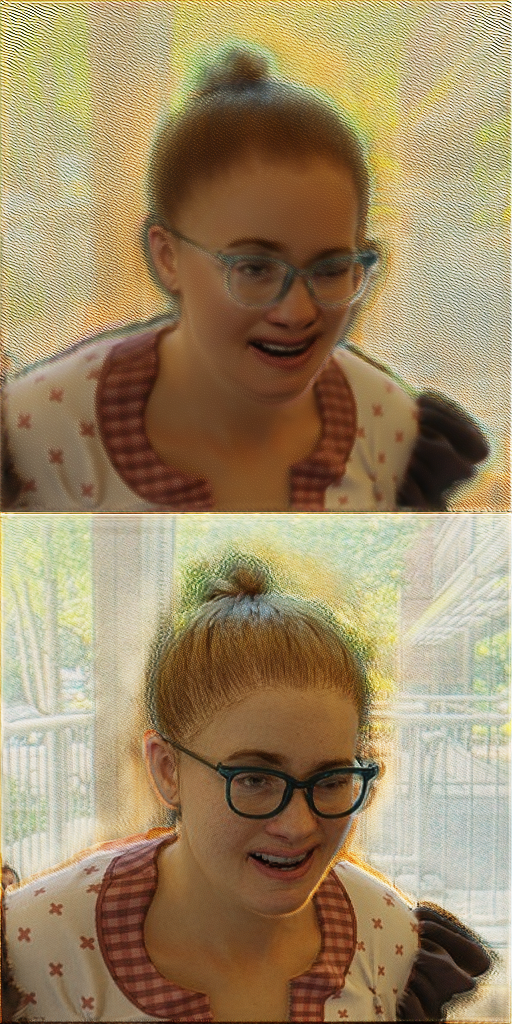}&
\includegraphics[height=\splatterfigheight]{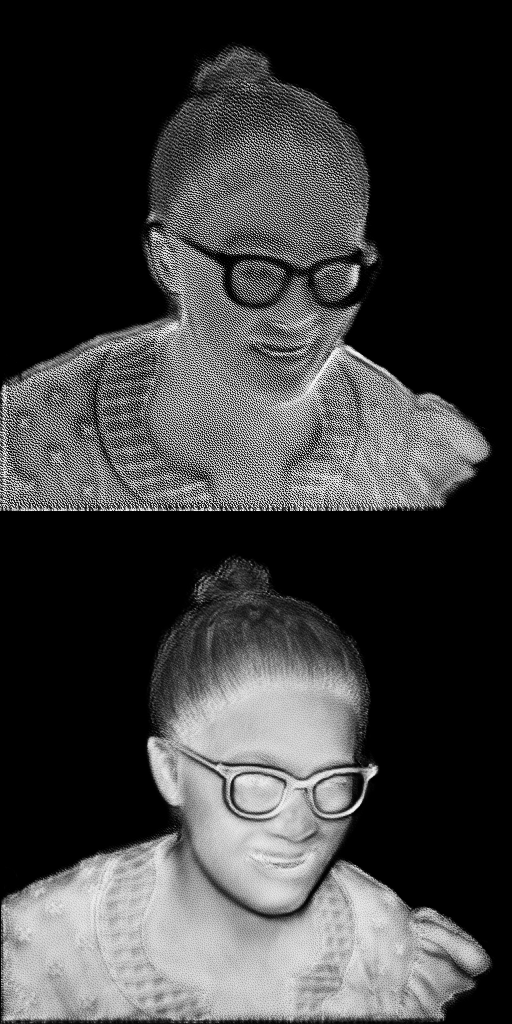}&
\includegraphics[height=\splatterfigheight]{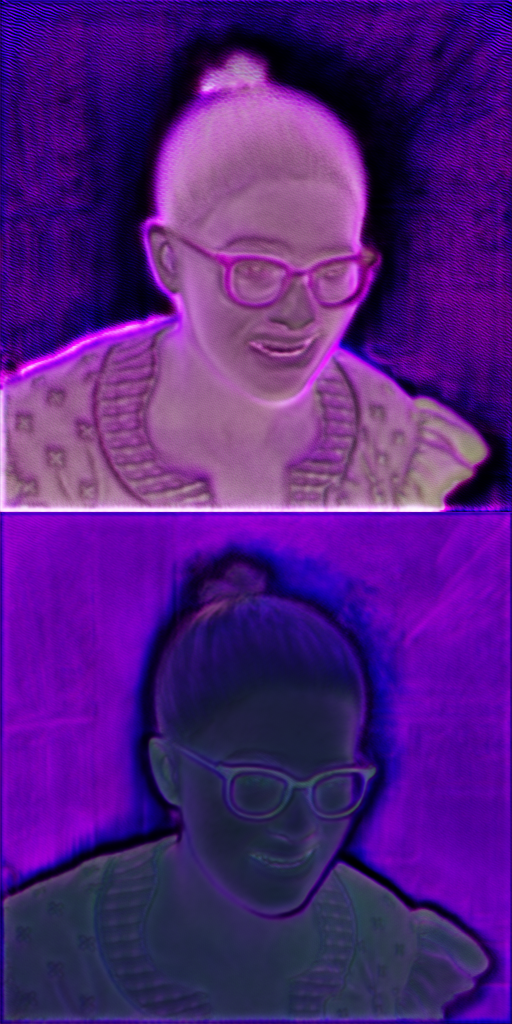}&
\includegraphics[height=\splatterfigheight]{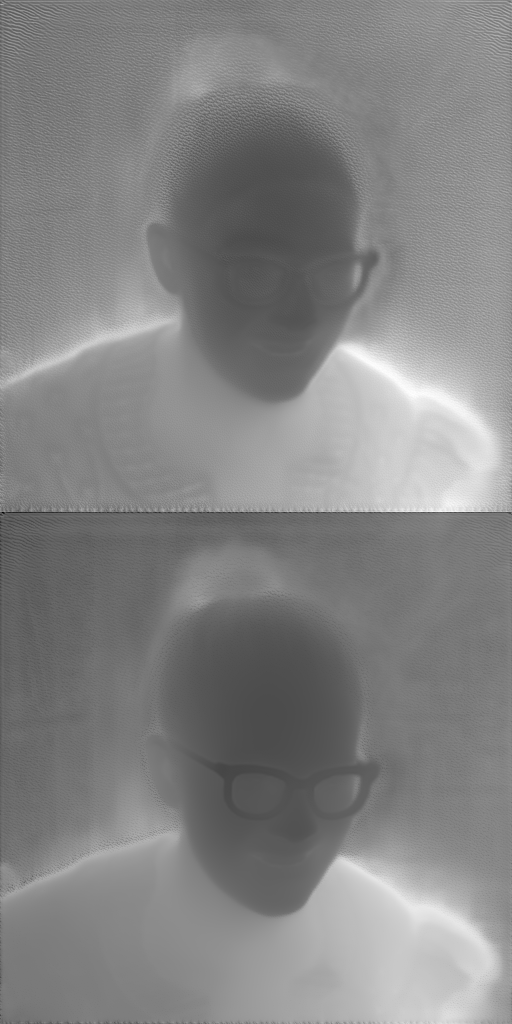}&
\includegraphics[height=\splatterfigheight]{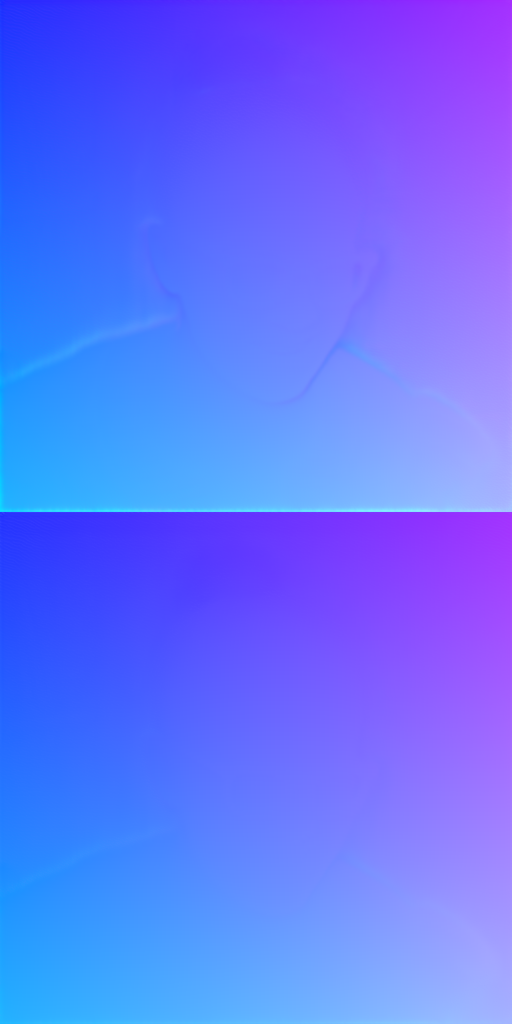}&
\includegraphics[height=\splatterfigheight]{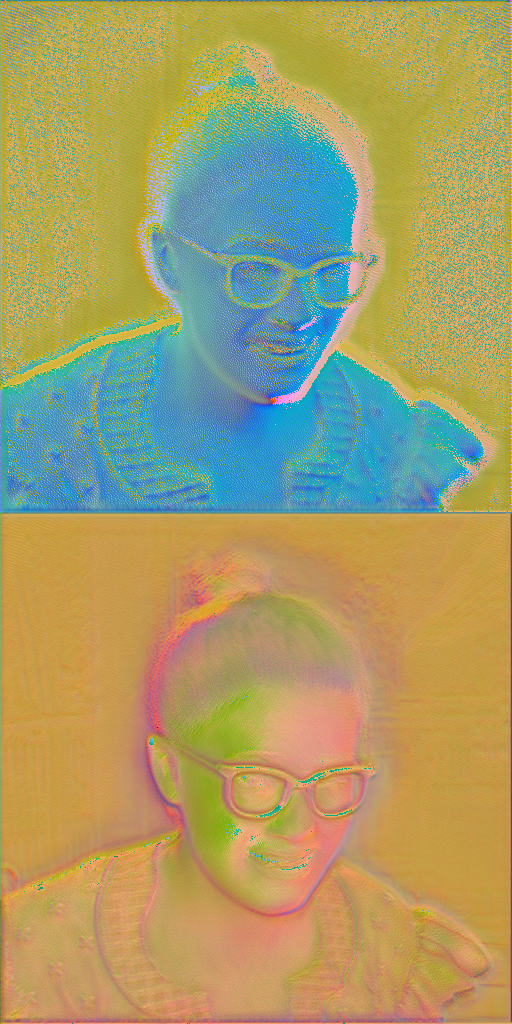}\\
(a)&(b)&(c)&(d)&(e)&(f)&(g)
\end{tabular}
    \caption{Splatter image components: a)~input ROI, b)~colour, c)~opacity, d)~log scale, e) z-depth, f) xyz, g) rotation (Euler angles)}
    \label{fig:splatter_image}
\end{figure}

\noindent\textbf{Optimizable layers.} The U-Net architecture uses only 2D convolutions and non-linear activation, which is limited in its ability to create output with high spatial frequencies.
This limitation becomes stronger as we decrease the size of the network when aiming for a network that runs in real time. 
To mitigate this problem, we add explicit channels to the network that can take a different value for each pixel and whose values are optimized alongside the other parameters of the network. % i.e. the 2D convolution kernel weights. 
The optimizable channels enable a lightweight shallow network to generate high frequency patterns with a minor increase in compute, and so improve the quality of the final reconstruction. 
We add four channels at each stage of the decoding part of the U-Net architecture. 

\noindent\textbf{Direct colour sampling.}
Following \citet{szymanowicz24splatter}, we use offsets that allow 3D Gaussians to move away from the rays associated to their pixel in the splatter image.
For a Gaussian that is visible with large offset, the information provided by the skip connection in the U-Net is not well collocated with the location that the Gaussian would reproject in the input image.
As a result, the information most useful for predicting the Gaussian colour is not provided by the skip connection and needs to be propagated through the deeper layers, which limits the ability to maintain texture details.
In order to mitigate this, we compute the 2D location of the Gaussian after applying offset and projection into the input image and then use direct bilinear sampling of the input image where the Gaussian reprojects.
We provide this extra information as additional channels to a shallow convolutional residual block that predicts the Gaussian colour, scales, and rotation.

\noindent\textbf{Multiple Gaussians per pixel.}
In order to provide more flexibility to the representation, we generate multiple Gaussians per pixel~\cite{szymanowicz2024flash3d}.
This allows the network to predict multiple layers of the surface while only increasing the computational cost of the network by a small amount.
We determined experimentally that using two Gaussians per pixel provided a good quality improvement, while three did not yield noticeable improvements and significantly increased rendering time. As visible in \cref{fig:splatter_image} the Gaussians in the first layer have larger scales and seem to contribute more to occluded regions, while the ones in the second layer have much smaller scales and tend to represent high-frequency details in the visible regions.

\noindent\textbf{ROI homography mapping.}
Our method runs on a region of interest (ROI) around the face, rather than the whole image.
The simplest approach to extract an ROI is a simple image crop.
However, this has important limitations when it comes to predicting 3D geometry; faces away from the image centre appear distorted in the cropped images due to the perspective projection.
CLIFF~\cite{CLIFFZhihao2022} addresses this by providing the focal length and central point of a new virtual camera with oblique frustum that corresponds to the cropped region as input to the DNN.
We instead create a virtual camera with a symmetric frustum, that is rotated to point to the centre of the face in the original image, with the FOV computed such that the face covers a third of the FOV angle. 
We then warp the input image to match this new camera using a homography. 
Because this camera has a symmetric frustum, we only need to provide the normalized focal length $f/w$  as input to the neural network. We concatenate the input image with the ray directions and two channels that contain the repeated normalized focal length and its inverse.
\label{sec:homography}

\subsection{Data generation}
Our method takes one image as input and generates a set of Gaussians that are rendered in new views using a differentiable renderer. 
These renders are compared to ground truth views to obtain a training loss.
For training we therefore need a multi-view dataset, but capturing a diverse, multi-view, in-the-wild dataset at a scale suitable to train a Gaussian splat prediction model is a formidable challenge. Instead, we use \emph{exclusively} synthetic data to train our method, employing the data generation pipeline described by \citet{hewitt2024look}.
This enables us to create multi-view data with perfect camera parameters, including diverse subjects, and emulating in-the-wild scenarios.
Our synthetic dataset includes $N_s = \text{20,000}$ subjects with varied face shape, skin texture, hair, clothing, glasses, body pose and facial expression. 

In order for the prediction to align well with the ground truth in the novel views we need the model to predict accurate absolute positions of the Gaussians.
Due to the depth ambiguity that is inherent to single-view reconstruction, this is a very challenging task.
To resolve this, previous Gaussian splat prediction methods train models using synthetic data with 3D objects positioned at fixed distance to the camera~\cite{szymanowicz24splatter}. 
This approach has critical limitations: the same face looks very different when captured from cameras at different distances; by training the model with a fixed distance, it cannot learn strong priors about the 3D proportions of faces.
Further, absolute depth prediction is not required for our video call application, where small offsets are acceptable as long as the face shape is predicted correctly. 
Requiring the trained model to predict very accurate distances during training is therefore wasteful in terms of model capacity. 

\begin{figure}
    \centering
    \footnotesize
 \begin{tabular}{@{}c@{}c@{}c@{}c@{}}
    \includegraphics[height=0.22\linewidth]{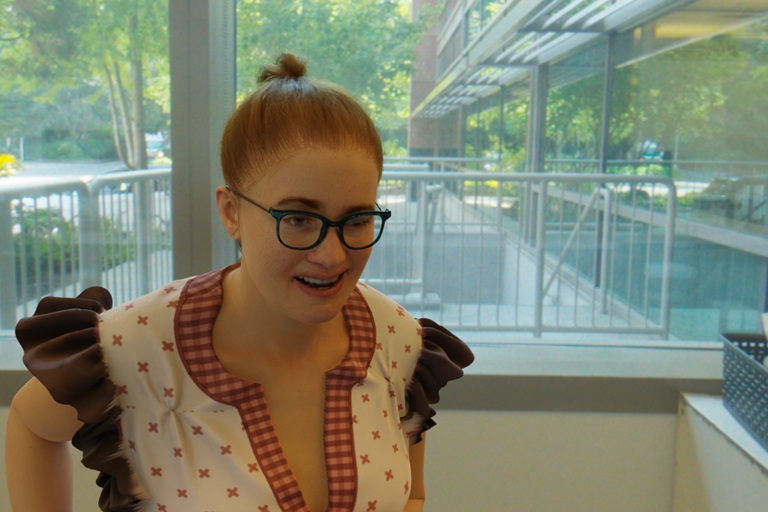}&
    \includegraphics[height=0.22\linewidth]{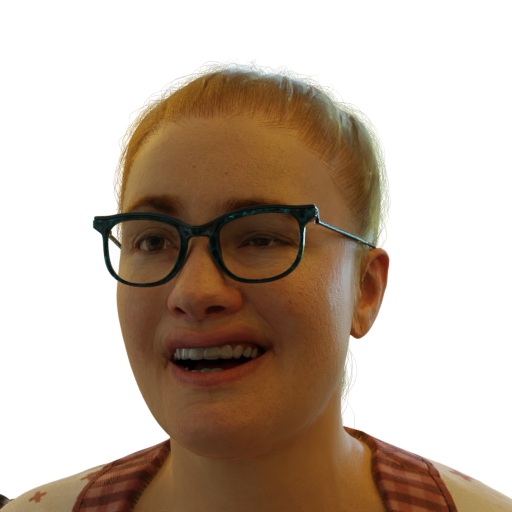}&
    \includegraphics[height=0.22\linewidth]{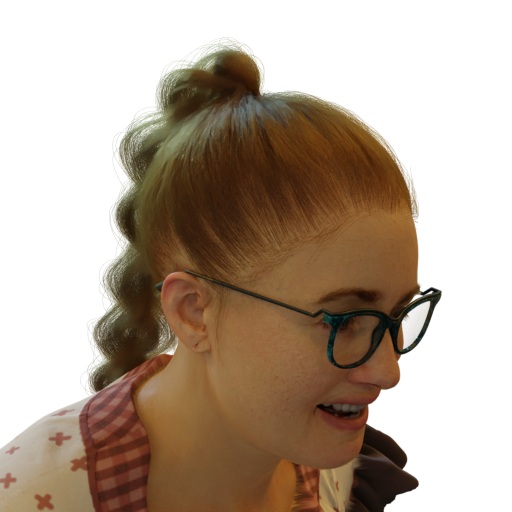}&
    \includegraphics[height=0.22\linewidth]{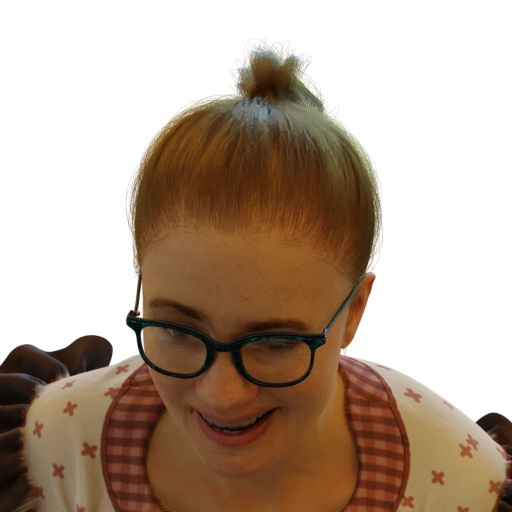}\\
    (a) & (b) & (c) & (d)
 \end{tabular}
    \caption{Example images from our synthetic training data. a) Input image with background. b-d) Three renders (out of 10) from which the foreground image and alpha mask are used for supervision. Note the presence of glasses and different hair styles.
    }
    \label{fig:training_data}
\end{figure}

We instead resolve the depth ambiguity by rescaling the prediction during training (see \cref{sec:training_loop}).
This is enabled by high-quality ground-truth depth generation from the synthetic pipeline.
As such, we sample input cameras at various distances to the subject (ranging from 40cm to 1m).
The input camera should emulate a webcam, so we position faces randomly within the image, ensuring that the angle of the face to the camera is within 30\textdegree{}, and render at 1080$\times$720 resolution.
The images used for the supervision do not need to reflect the variability in the input data. 
We therefore render 512$\times$512 pixel images from ten cameras closer to the face, pointing directly at the centre of the face, and whose positions are distributed randomly on a 45\textdegree{} spherical cap centred around the input camera direction.
Samples of our synthetic training data are shown in \cref{fig:training_data}.

\subsection{Training loop}

\begin{figure*}
    \centering
    \includegraphics[width=\linewidth]{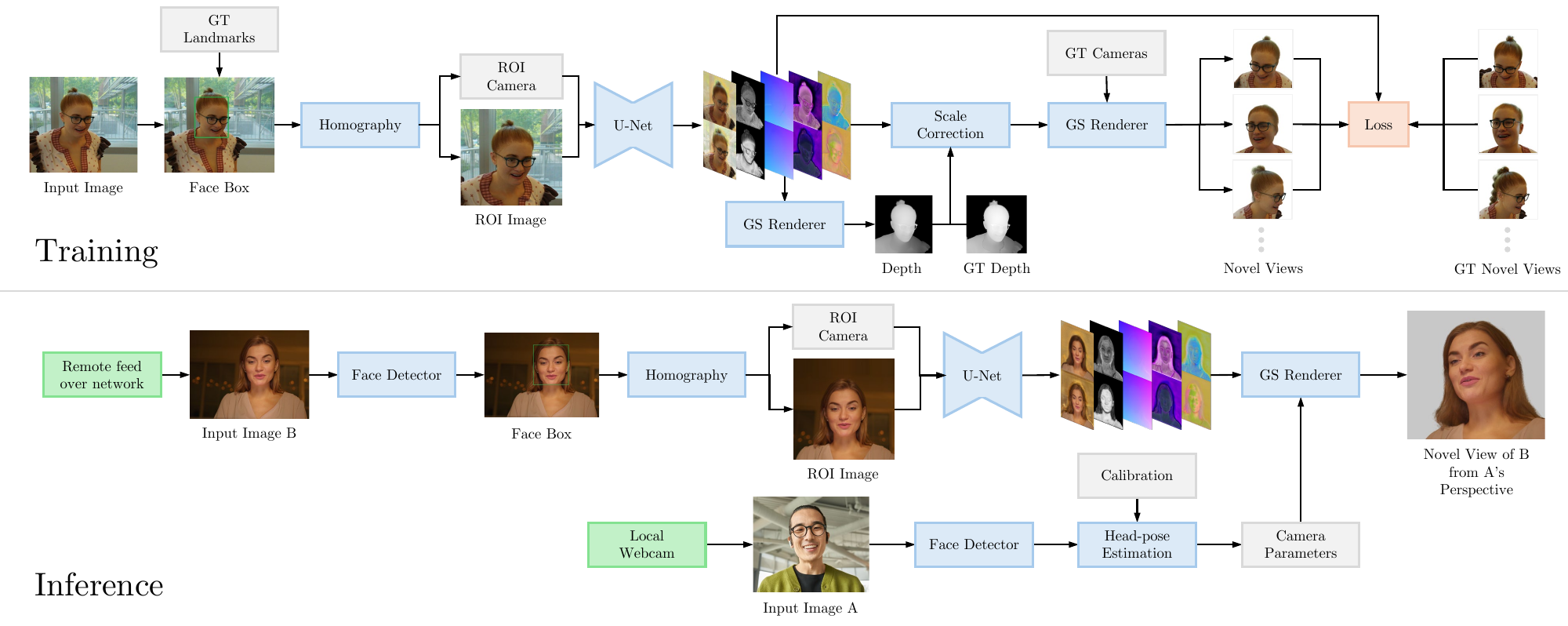}
    \caption{Overview of our the training (top) and inference (bottom) pipelines of our method.}
    \label{fig:pipeline}
\end{figure*}

\label{sec:training_loop}
The set of operations done on a single multi-view image sample in a batch during the training can be summarized as follows: (1) Using an ROI around the face based on the ground-truth 2D landmarks, define the rotated camera and warp the image using the homography as described in \cref{sec:homography}. Provide the warped image and ROI camera parameters as input to the DNN.
(2) Predict the splatter image in the ROI camera coordinate system.
(3) Render the splatter image Gaussian depth map into the input view using alpha composition to get differentiable rendering and gradients. Compute a global scale correction transform centred on the camera optical centre applied to the Gaussians by comparing the depth in the overlapping region of the ground truth mask and the rendered depth mask. The scale factor is obtained through linear least squares. 
(4) Apply the scale and transform the Gaussians back to the original camera coordinate system, render the corrected Gaussians in the virtual views and compare to ground truth images.
The process is visualized in the top of \cref{fig:pipeline}.

During training we minimize the following loss function
\begin{equation}
    L = L_d + \lambda_\sigma L_\sigma + \lambda_m L_m + \lambda_c L_c + \lambda_j L_j
\end{equation}
with data term $L_d$, regularizers $L_\sigma$, $L_m$ and $L_c$ and stability term $L_j$ and corresponding weights defined below.

Following \citet{szymanowicz24splatter} we define $L_d$ as:
\begin{equation}
L_d= \frac{1}{N_vN_s}\sum_{v=1}^{N_v}\sum_{s=1}^{N_s}L_i(R(U(I_{1}^s), C_v), I_v^s)
\end{equation}
where $L_i$ is an image pair loss,  $U$ is the U-Net model and $R$ is the differentiable Gaussian splat rendering function~\cite{kerbl3Dgaussians}.
$N_v=11$ is the number of views used in each sample for the supervision and $N_s$ is the number of multi-view samples in the dataset. $C_v$ are the camera parameters for view $v$ and $I_v^s$ is the ROI image for view $v$ in sample $s$.
$v=1$ is taken as the input view.
The image comparison loss, $L_i(X, Y)= L_e(X,Y) + \lambda_pL_p(X, Y)$, is composed of two terms: a Euclidean RGB distance term $L_e$ and a perceptual loss term $L_p$ with $\lambda_p=0.5$.
We composite the rendered image and the ground truth RGBA image over a random background colour for each sample. 

This encourages the network to perform alpha matting by estimating the Gaussian transparency and subtracting the background colour from the predicted Gaussian colours.

\noindent\textbf{Euclidean RGB distance.} We compute $L_e$ as the mean Euclidean distance in the the RGB space, i.e., 
$L_e(X, Y)=\frac{1}{HW}\sum_{i=1}^H\sum_{j=1}^W||X_{i,j}-Y_{i,j}||_2
$. This loss differs from the classical mean of absolute differences by being invariant to rotations in the 3D RGB colour space which helps reduce chromaticity aberration in the reconstruction, and is more tolerant to outliers than $L_2$ loss which helps to not over-penalize small silhouette misalignments.

\noindent\textbf{Perceptual loss.} We compute the perceptual loss $L_p$ using the VGG network~\cite{simonyan2014very} at two resolutions: full (512$\times$512) and half (256$\times$256). 
We empirically found that the perceptual loss at half resolution helps to obtain better results qualitatively and quantitatively.

\noindent\textbf{Regularization Losses.} 
We extend the original Splatter Image~\cite{szymanowicz24splatter} method by predicting multiple Gaussians per pixel.
With this modification we often encounter local minima during training where only one `layer' of Gaussians is used, even though this does not provide the best results.
We therefore introduce losses on the splatter image opacities to avoid falling into poor local minima by penalizing cases where the average opacity is near 0 in any of the layers.
Denoting $\sigma_{i,j,k}$ the opacity of the $k^{\text{th}}$ Gaussian represented at coordinate $(i, j)$, we compute mean opacities for Gaussians layer $k$ as $\bar{\sigma}_{k}=\frac{1}{HW} \sum_{i,j}\sigma_{i,j,k}$. We penalize the means $\bar{\sigma}_{k}$ near zero using a loss $L_m=\frac{1}{K}\sum_{k}e^{-\tau\bar{\sigma}_{k}}$ with weight $\lambda_m=5$ and $\tau=50$, where $K$ is the number of Gaussians per pixel.
We also encourage the prediction of opaque Gaussians by using an opacity loss $L_\sigma=\frac{1}{K}\sum_{k}(1-\bar{\sigma}_{k})$
with weight $\lambda_\sigma=10^{-4}$. 
This term acts as a small bias and helps in making the best use of the allowed number of Gaussians.
Finally, we add a regularization term $L_c=\log(s)^2$ on the scale correction, $s$, obtained in the third step of the training loop to encourage the network to predict valid metric depths, with weight $\lambda_c=1$.
This helps significantly with convergence during training.

\noindent\textbf{Jitter Loss.}
Temporal stability is important in the context of videoconferencing; flicker can be very distracting and degrade the experience. 
In order to improve the temporal stability of our prediction we use an approach similar to \citet{ZhengSLG16StabilityTraining}. 
Each sample, $s$, in a training batch is duplicated and random perturbations added to the face box, independently for each sample, to form sample $s'$.
We then encourage stability by penalizing the difference in the renders between each pair of duplicated samples, $L_j = \frac{1}{N_vN_s}\sum_{v=1}^{N_v}\sum_{s=1}^{N_s}\|R(U(I_1^{s}), C_v)-R(U(I_1^{s'}), C_v)\|^2$.
This encourages the DNN to favour consistency in the prediction over accuracy, and we can control the trade-off with weight $\lambda_j = 1.0$.

\section{3D Video Calls on Commodity Hardware}
\label{sec:runtime}

\begin{figure}
    \centering
    \includegraphics[width=\linewidth]{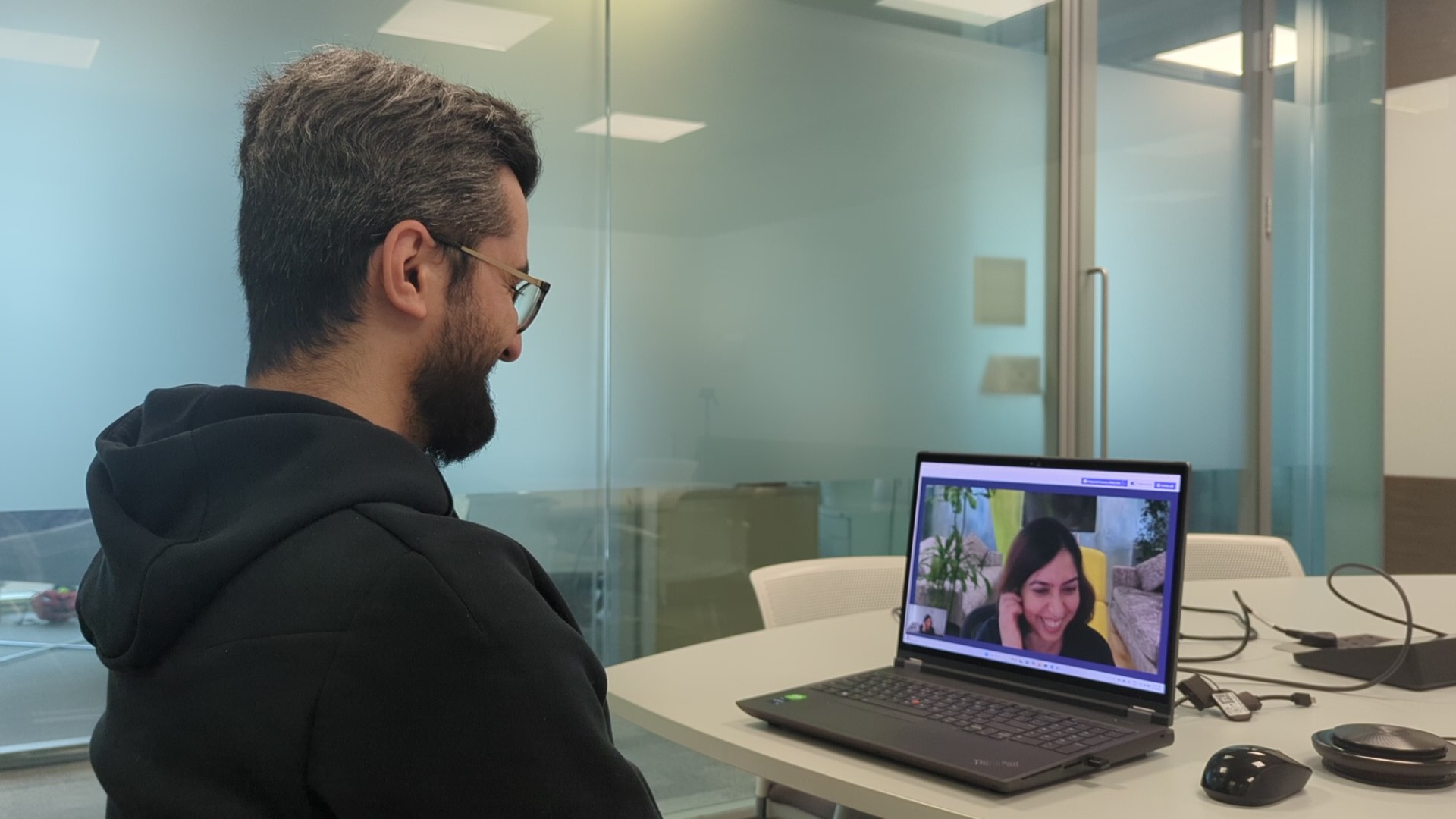}
    \caption{Snapshot from our prototype 3D video conferencing application. See supplementary video for details.}
    \label{fig:live-demo}
\end{figure}
This section describes the live 3D video conferencing prototype application we use to demonstrate our method. 
Our application exploits the `magic window' effect~\cite{lee2007magicwindow} to create the illusion of 3D on a 2D display.
We have two subjects, A and B, both equipped with a webcam, a 2D computer screen and a GPU. 
Given the symmetry of the setup, we describe only what happens on subject A's machine, the bottom of \cref{fig:pipeline} provides a graphical representation.
The camera-screen relative pose is assumed known through calibration. 

We render the appropriate new view of B's face on A's screen through the steps: (1) Detect A's face using an SSD-based detector~\cite{liu2016ssd} to get a bounding box, smooth it temporally to reduce jitter, and estimate the position of A's head assuming a fixed distance to the camera.
(2) Detect B's face in the video feed received from B, and extract the ROI around B's face using a homography, as described in \cref{sec:homography}.
(3) Predict a Gaussian splatter image in the ROI camera using the model described in \cref{sec:training_loop}.
(4) Transform the Gaussians back to the original input camera coordinate system for B's video and render the 3D reconstruction of B in the view of subject A based on their estimated head position.

We run the prototype on a laptop with a NVIDIA 4090 Mobile GPU and video sent over a standard network connection.
The application runs at 28FPS, a snapshot of the prototype in action is shown in \cref{fig:live-demo}, further demonstration is provided in the supplementary video.

\section{Experiments and Results}

To validate our approach, we compare our method to recent works across a number of benchmarks, for ablation experiments please consult the supplementary material.
Note that we limit comparison to reconstruction methods which satisfy our requirement of \emph{authenticity}, i.e., those which do not rely on an enrolled representation.

\subsection{Experimental Setup}
\newcolumntype{C}{>{\centering\arraybackslash}X}
\newcommand{\cafcawidth}{0.1666\linewidth}
\begin{figure}
    \centering
    \scriptsize
    \begin{tabularx}{\linewidth}{@{}C@{}C@{}C@{}C@{}C@{}C@{}C@{}C@{}}
         \multirow{2}{*}{Input} & FaceLift & TriPlaneNet & SI & \multirow{2}{*}{Ours} & Ground \\
         & \cite{lyu2024faceliftsingleimage3d} & \cite{bhattarai2024triplanenet} & \cite{szymanowicz24splatter} & & Truth \\
    \end{tabularx}  
    \includegraphics[width=\cafcawidth]{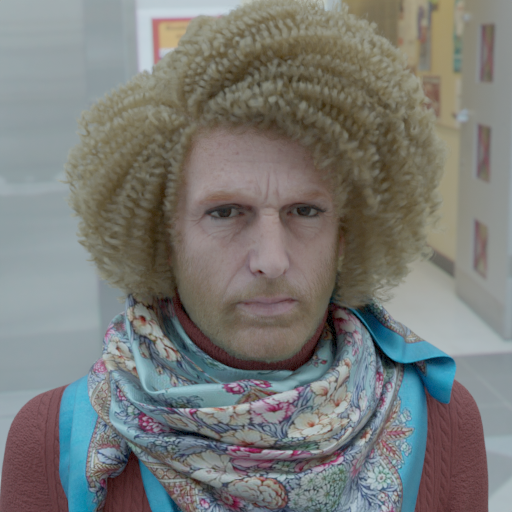}\includegraphics[width=\cafcawidth]{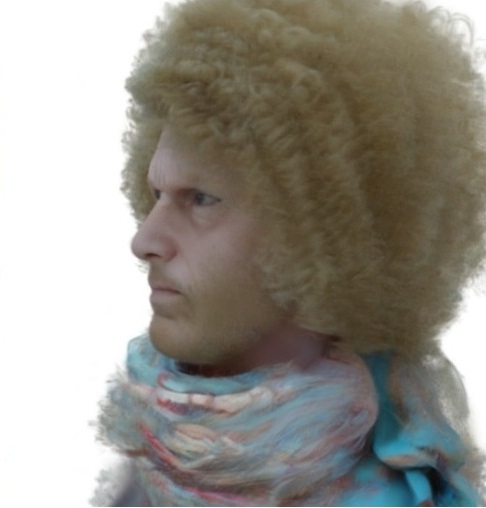}\includegraphics[width=\cafcawidth]{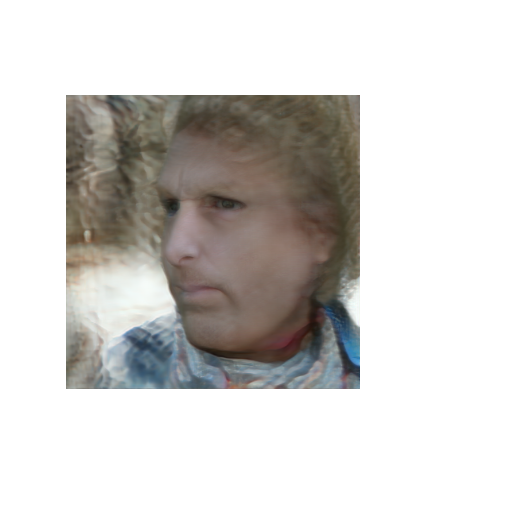}\includegraphics[width=\cafcawidth]{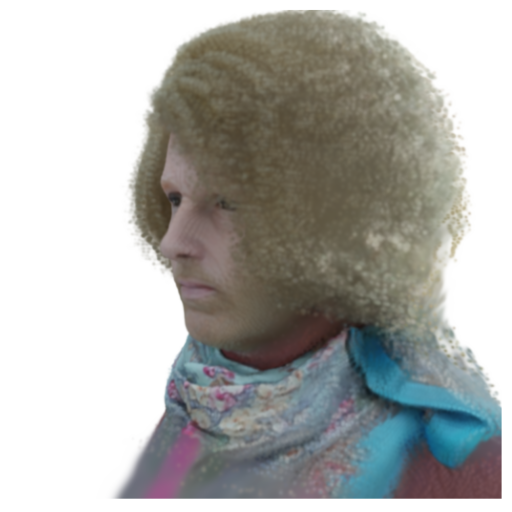}\includegraphics[width=\cafcawidth]{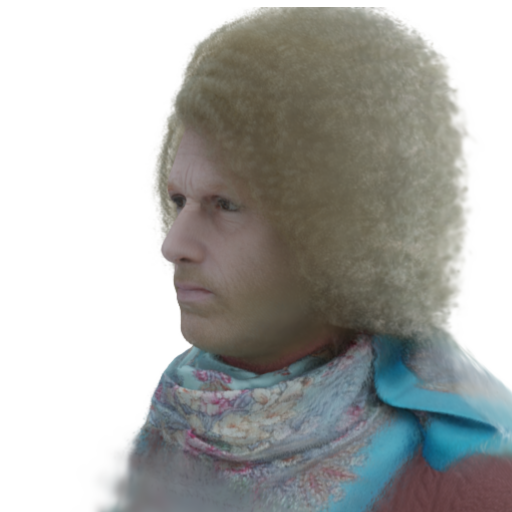}\includegraphics[width=\cafcawidth]{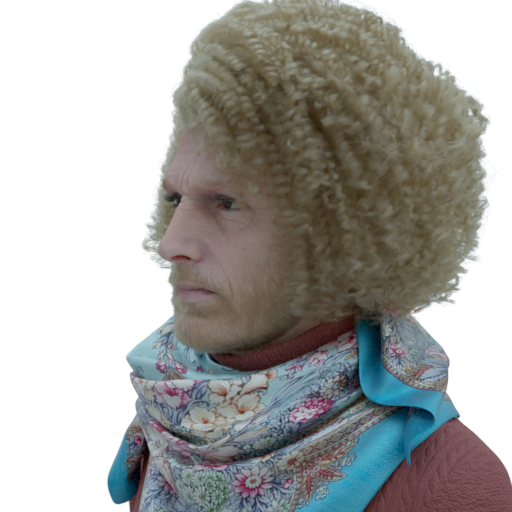}\\
    \includegraphics[width=\cafcawidth]{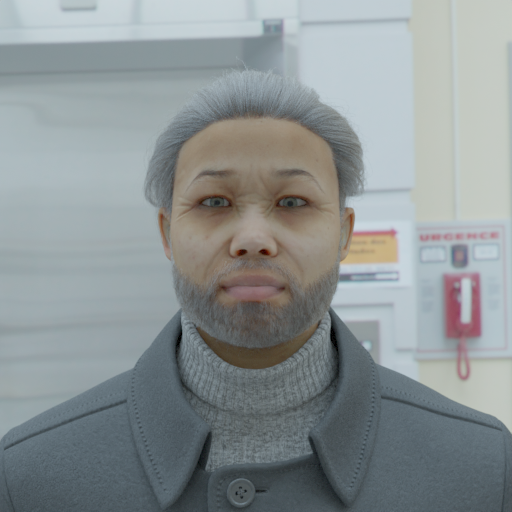}\includegraphics[width=\cafcawidth]{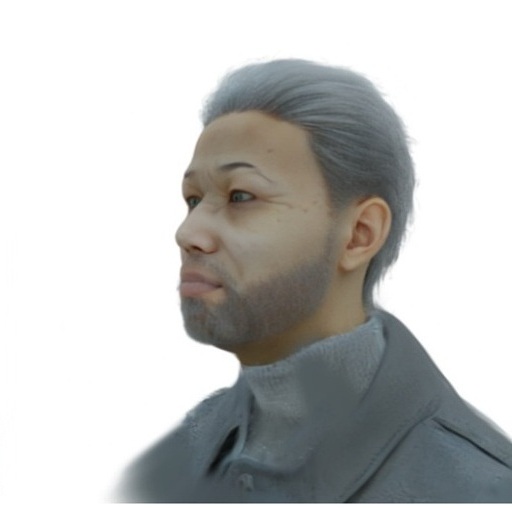}\includegraphics[width=\cafcawidth]{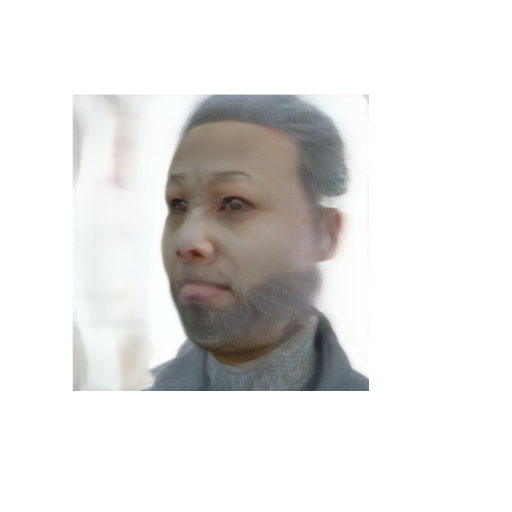}\includegraphics[width=\cafcawidth]{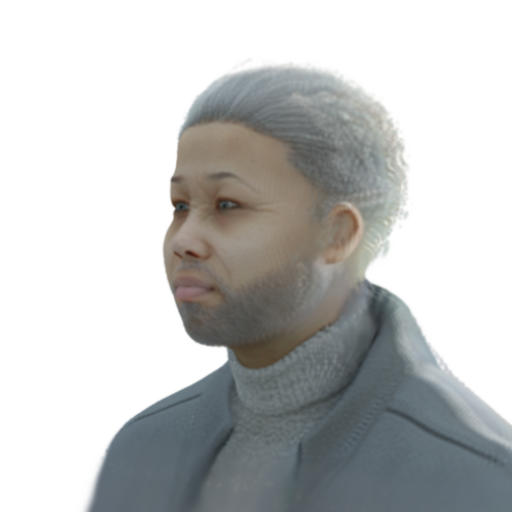}\includegraphics[width=\cafcawidth]{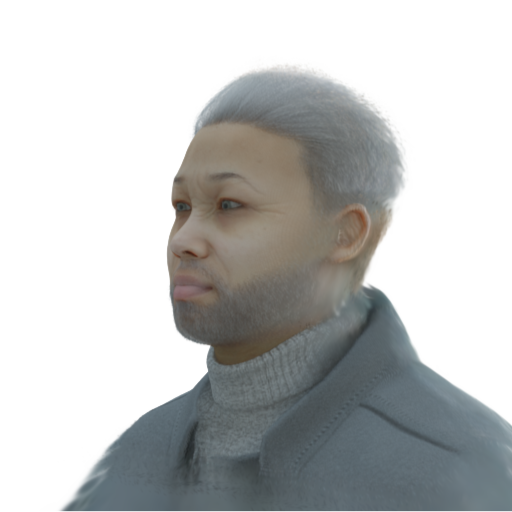}\includegraphics[width=\cafcawidth]{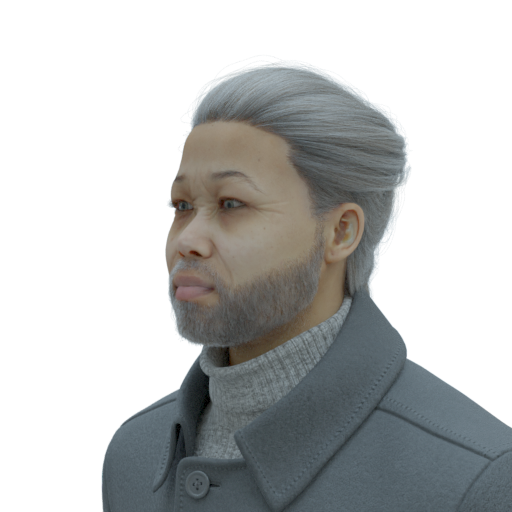}\\
    \caption{Qualitative comparison for images from the Cafca dataset \cite{buehler2024cafca}. Splatter Image (SI) has been retrained on our data.}
    \label{fig:qual_cafca}
\end{figure}
\newcolumntype{s}{>{\hsize=1.8cm \centering}X}
\newcommand{\avafigheight}{2.76cm}
\begin{figure}
    \centering%
    \scriptsize%
    \begin{tabularx}{\linewidth}{@{}s@{}C@{}C@{}C@{}C@{}C@{}C@{}}
         \multirow{2}{*}{Input} & FaceLift &TriPlaneNet&SI & \multirow{2}{*}{Ours} & Ground \\
         & \cite{bhattarai2024triplanenet}
         &\cite{lyu2024faceliftsingleimage3d} &\cite{szymanowicz24splatter}& & Truth
    \end{tabularx}
\includegraphics[width=1.7cm]{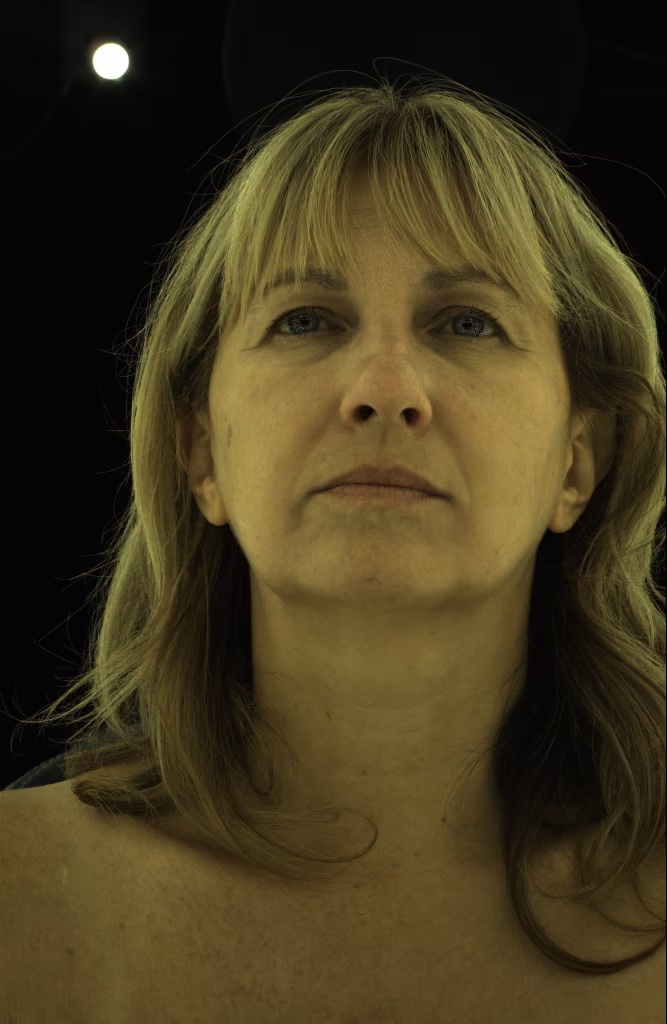}%
\includegraphics[height=\avafigheight]{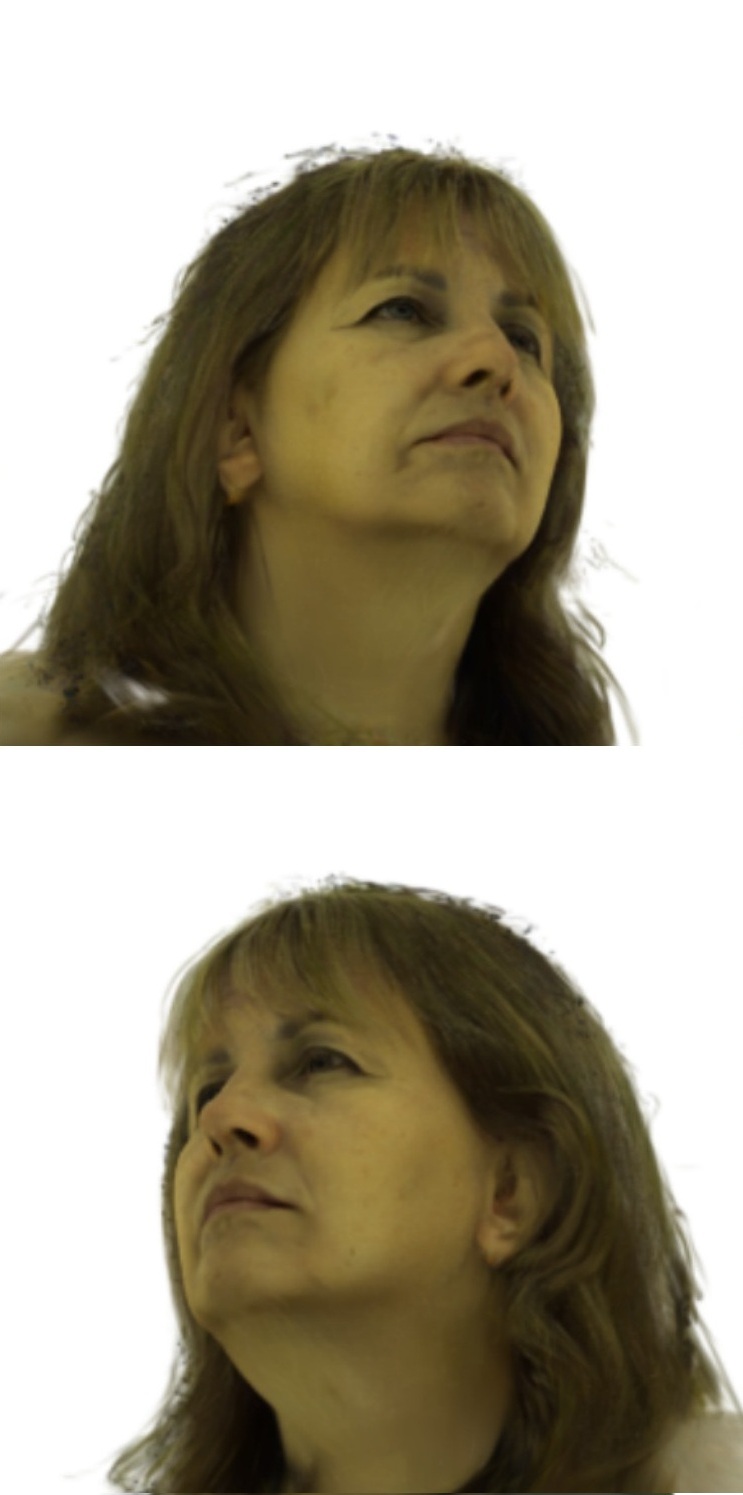}%
\includegraphics[height=\avafigheight]{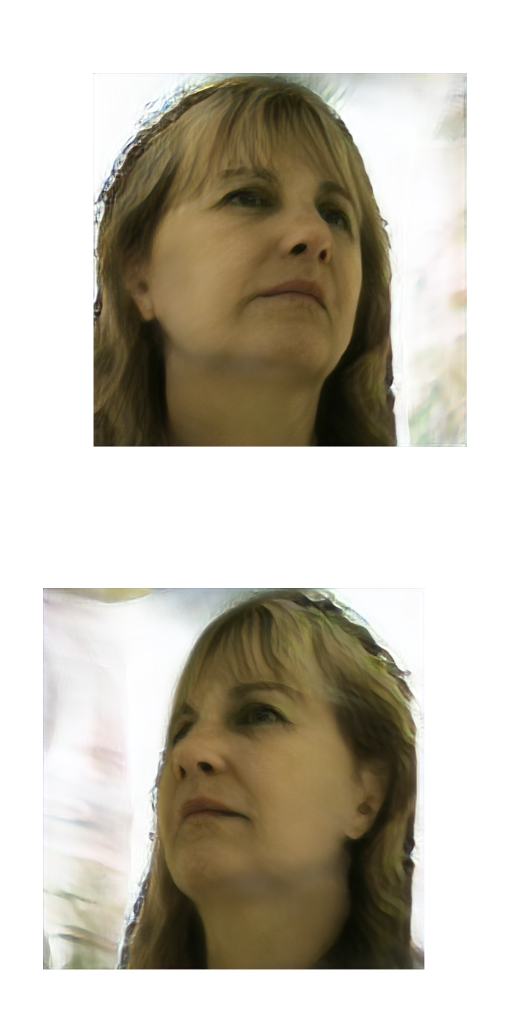}%
\includegraphics[height=\avafigheight]{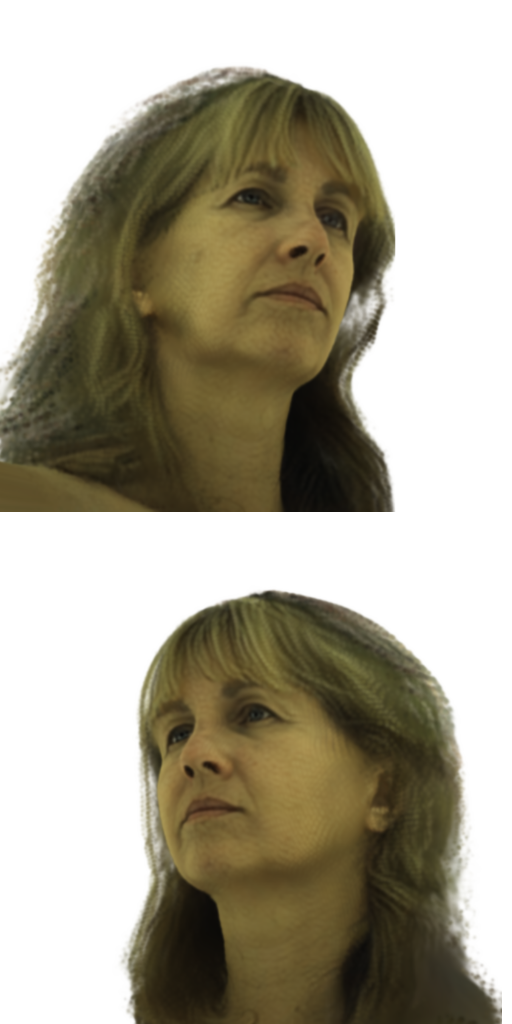}%
\includegraphics[height=\avafigheight]{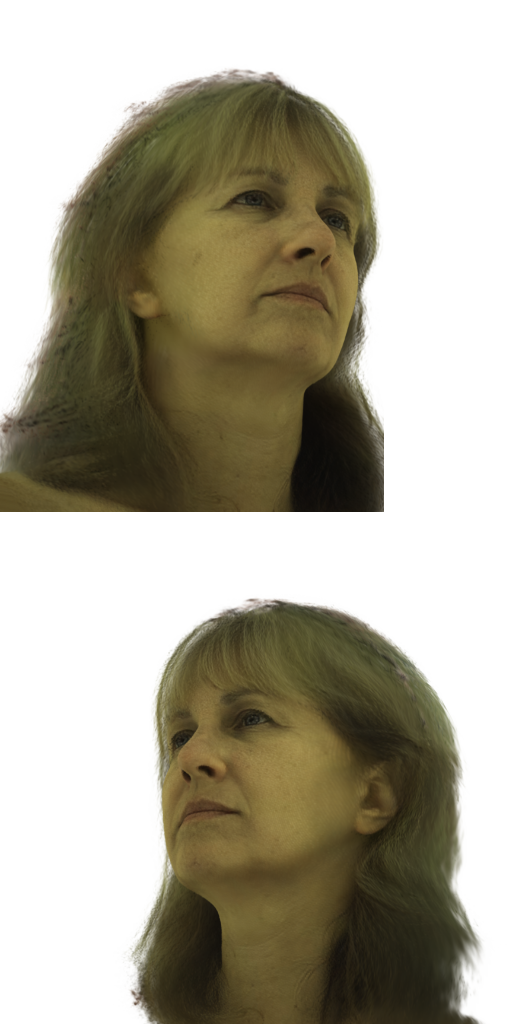}%
\includegraphics[height=\avafigheight]{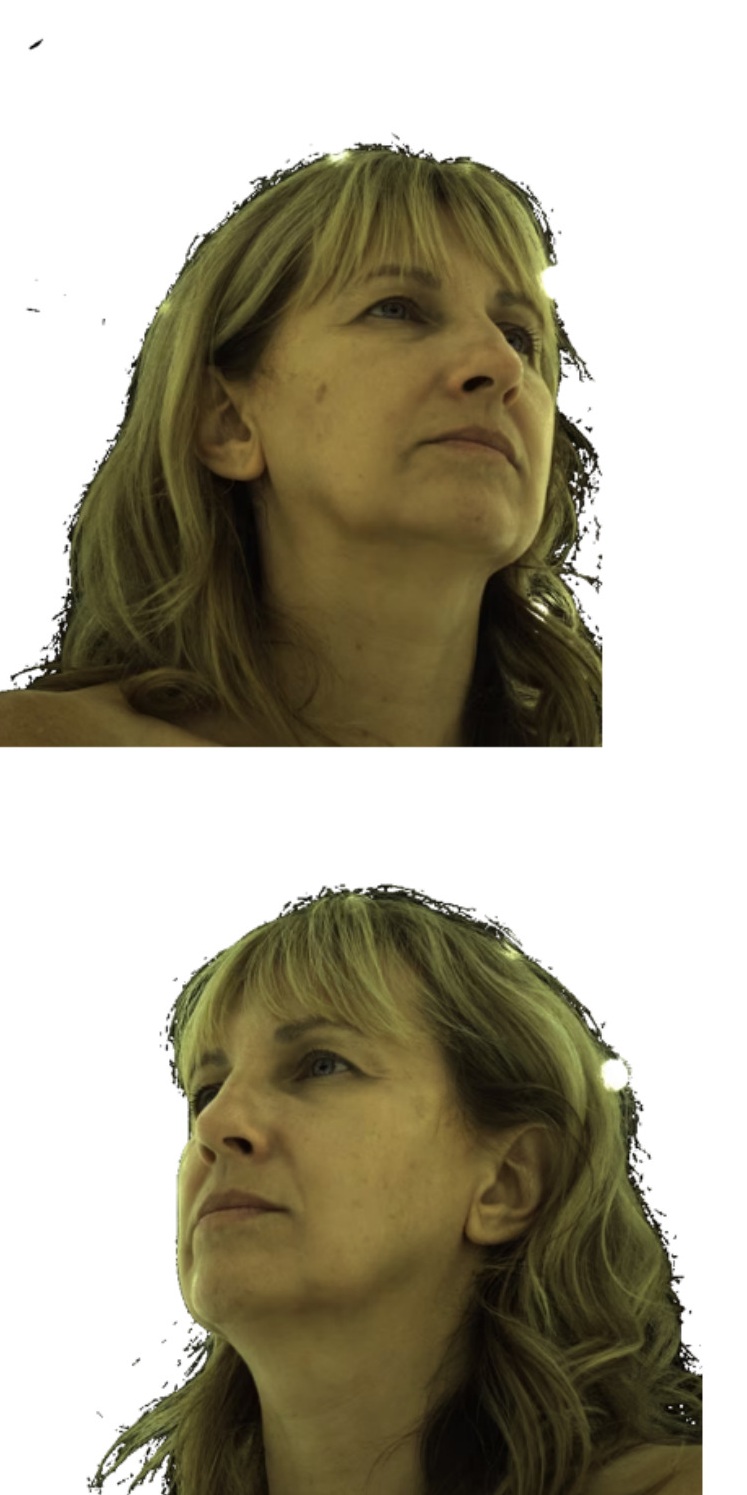}
\includegraphics[width=1.7cm]{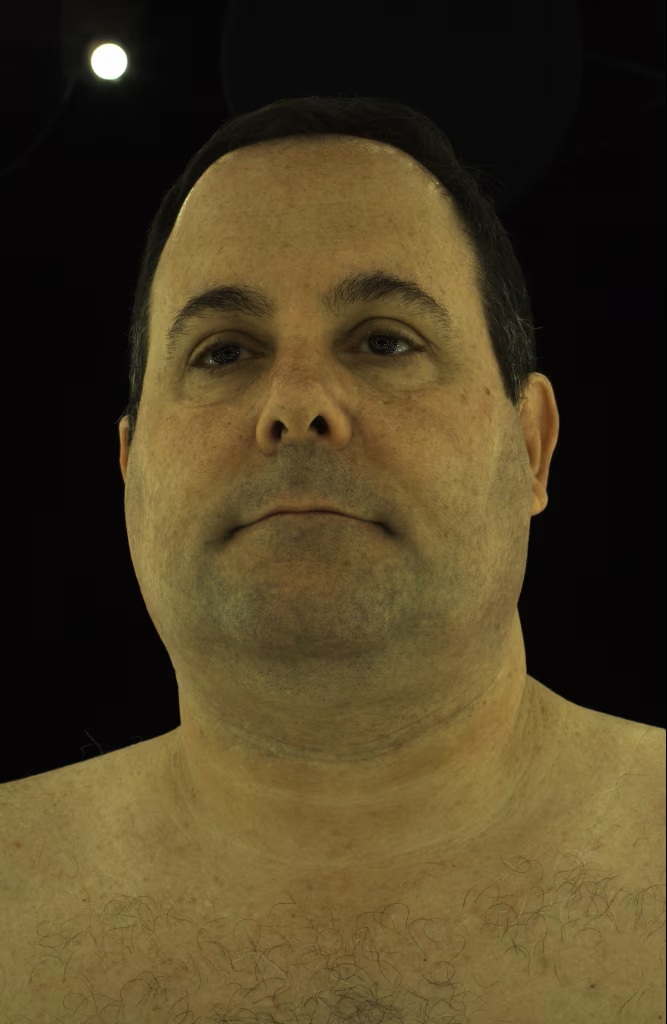}%
\includegraphics[height=\avafigheight]{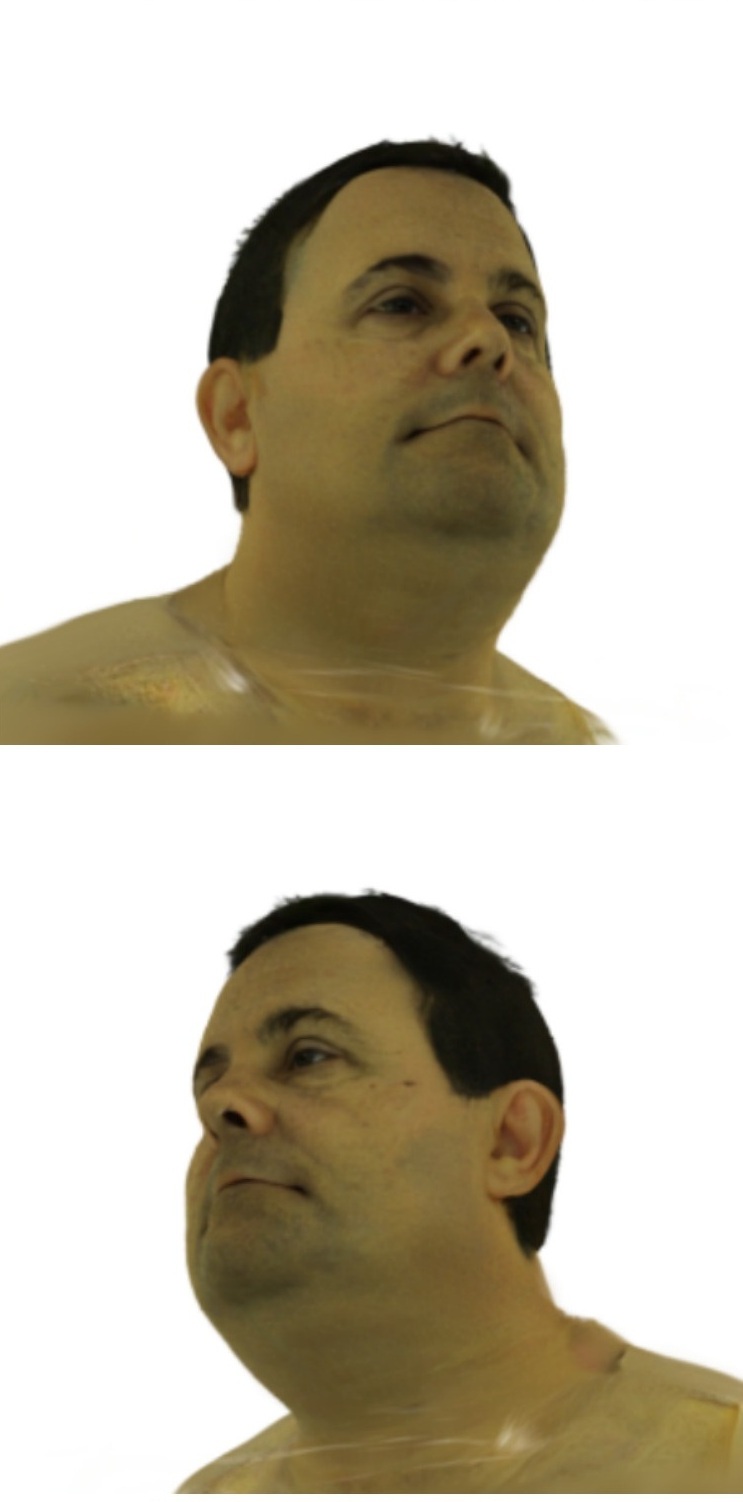}%
\includegraphics[height=\avafigheight]{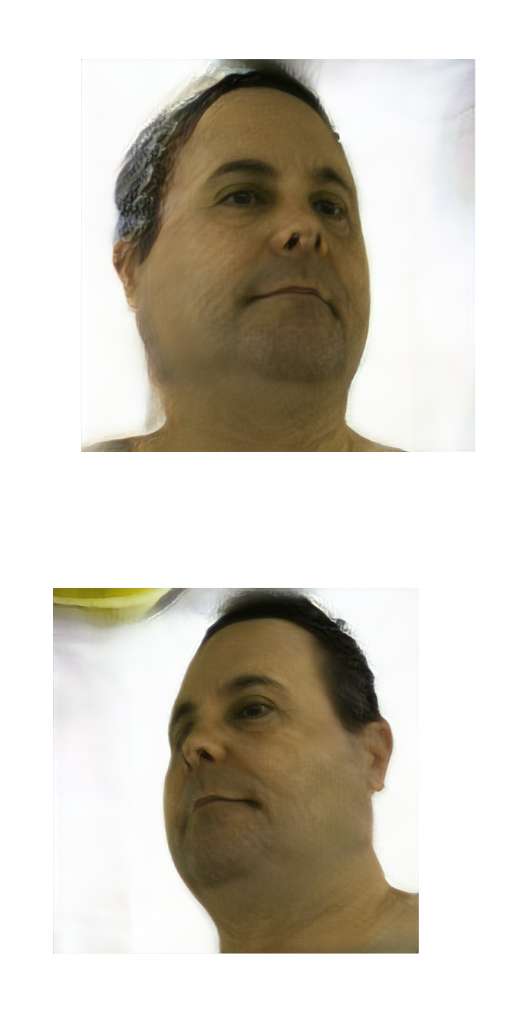}%
\includegraphics[height=\avafigheight]{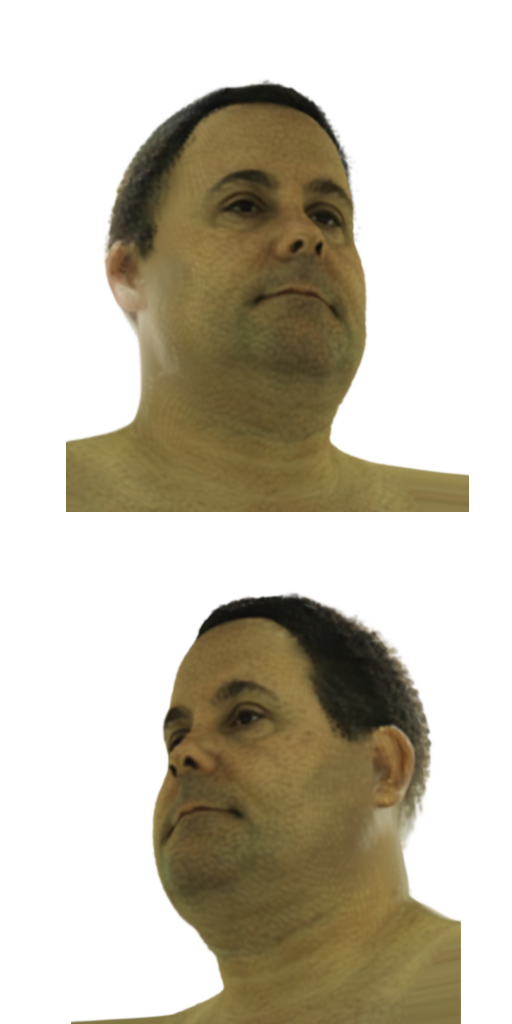}%
\includegraphics[height=\avafigheight]{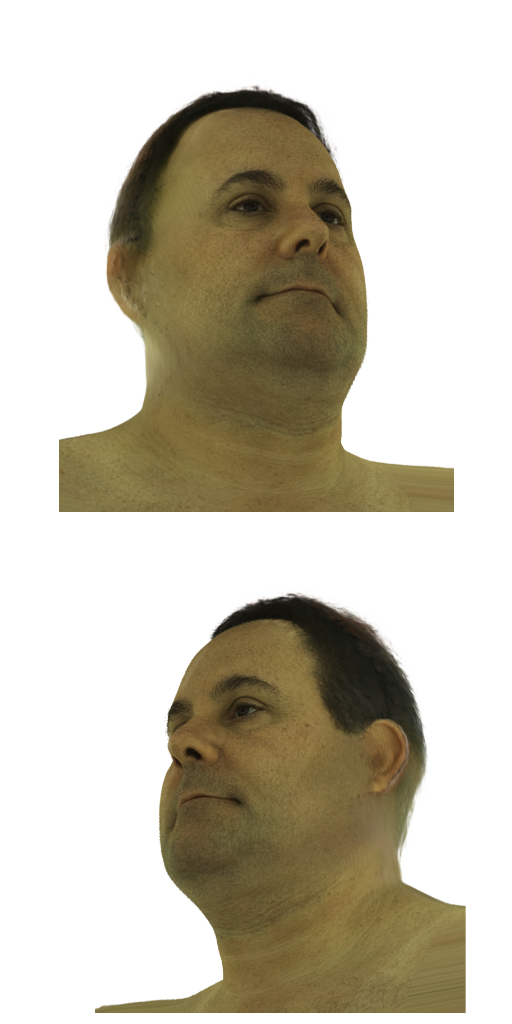}%
\includegraphics[height=\avafigheight]{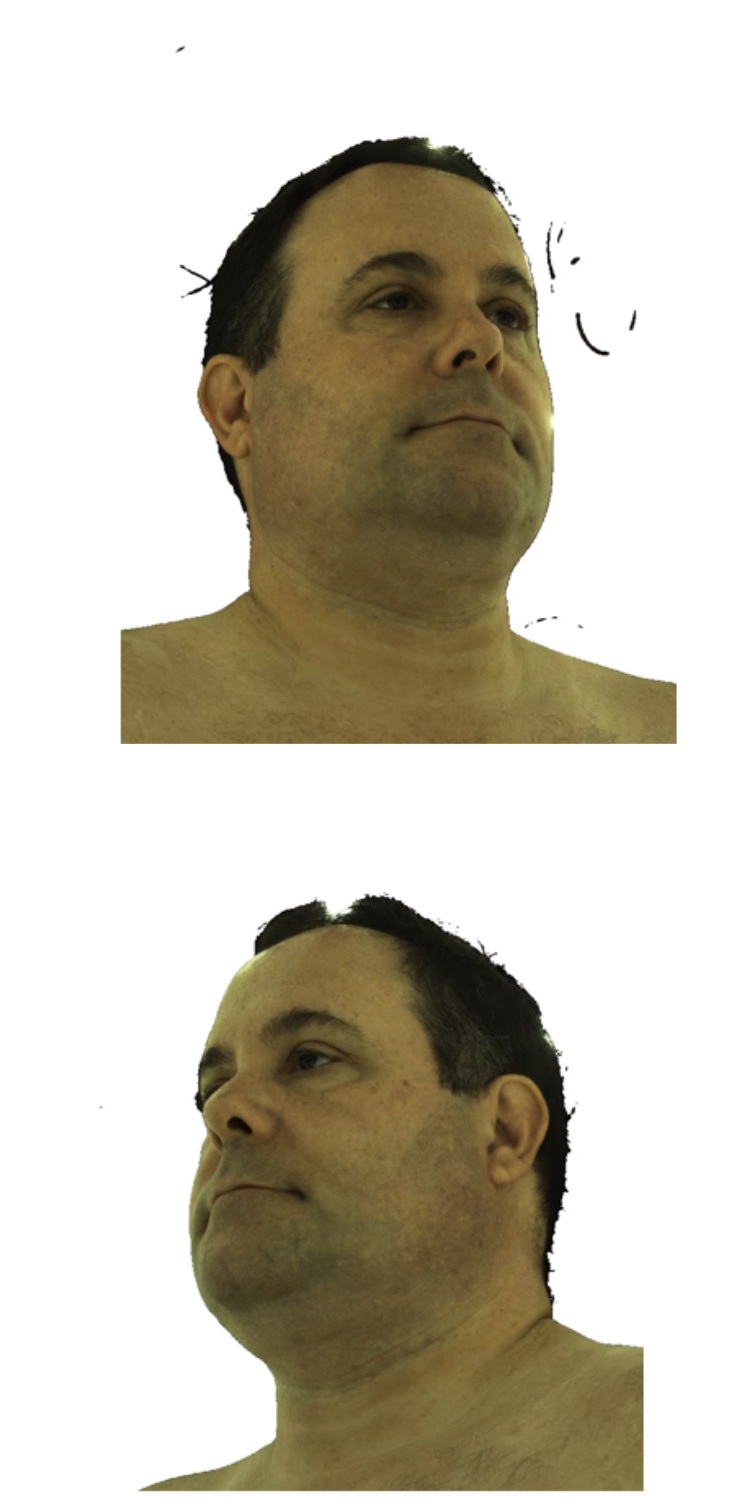}
    \caption{Qualitative comparison for images from the Ava-256 dataset \cite{martinez2024codec}. Splatter Image (SI) has been retrained on our data.}
    \label{fig:qual_ava256}
\end{figure}
\begin{figure}
    \centering
    \scriptsize
    \begin{tabularx}{\linewidth}{CCCCC}
        \multirow{2}{*}{Input} & \multicolumn{2}{c}{Live 3D Portrait~\cite{trevithick2023}} & \multicolumn{2}{c}{Ours} \\
        & Image & Geo & Image & Geo \\
    \end{tabularx}
        \includegraphics[width=0.2\linewidth]{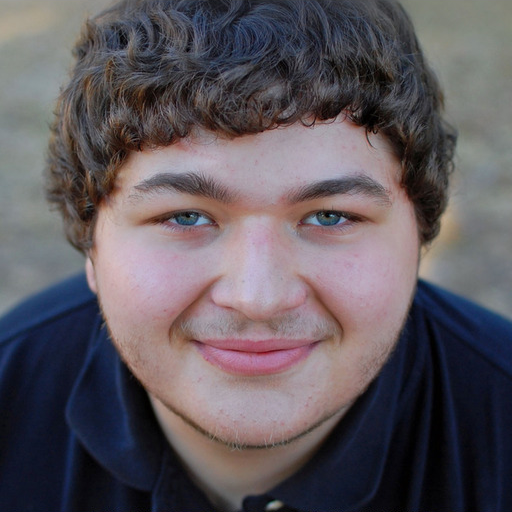}\includegraphics[width=0.2\linewidth]{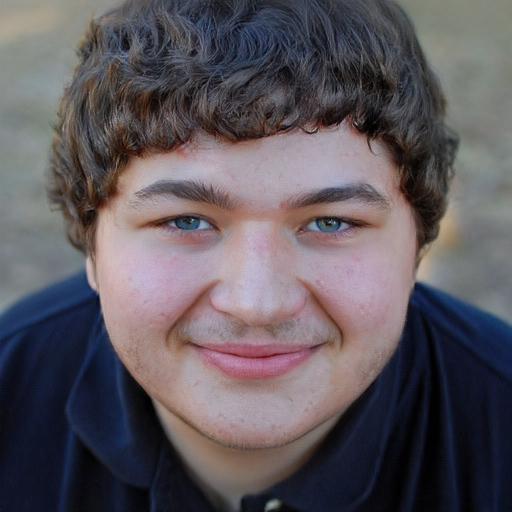}\includegraphics[width=0.2\linewidth]{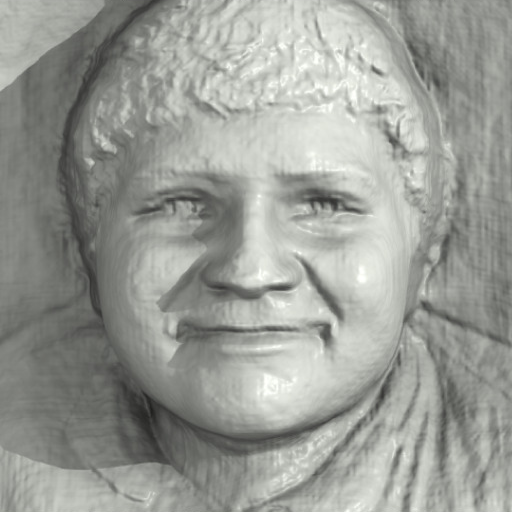}\includegraphics[width=0.2\linewidth]{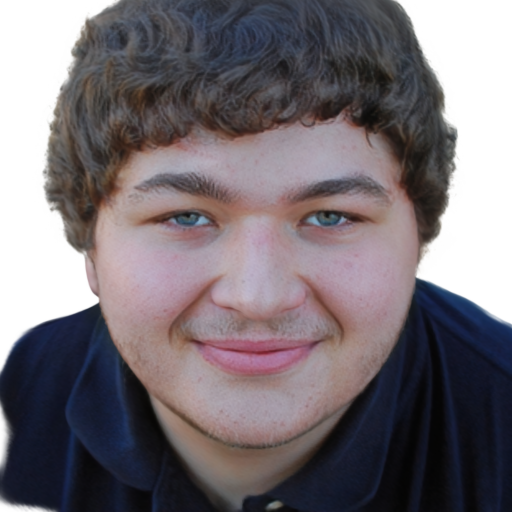}\includegraphics[width=0.2\linewidth]{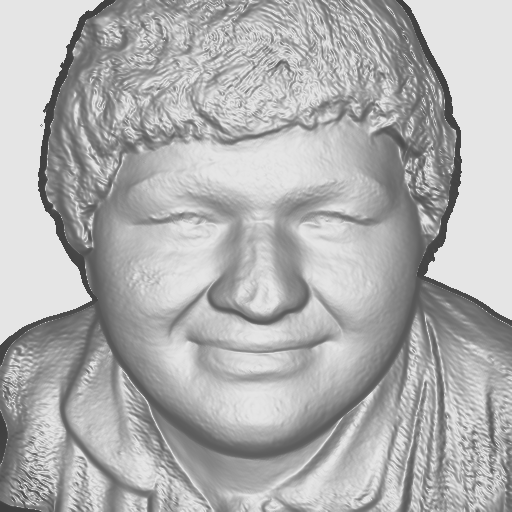} \\
        \includegraphics[width=0.2\linewidth]{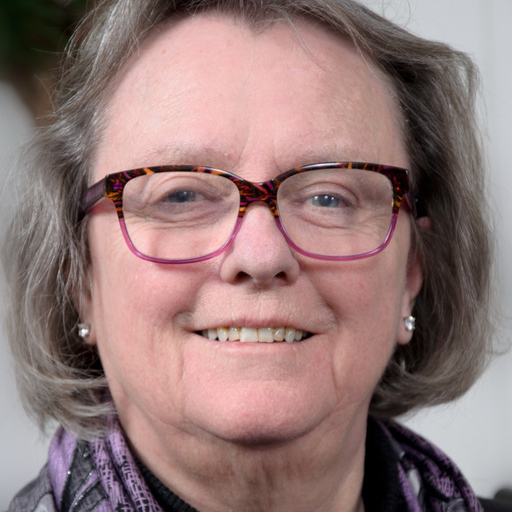}\includegraphics[width=0.2\linewidth]{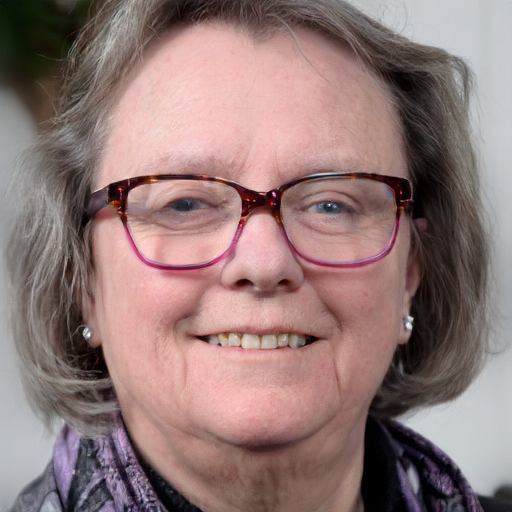}\includegraphics[width=0.2\linewidth]{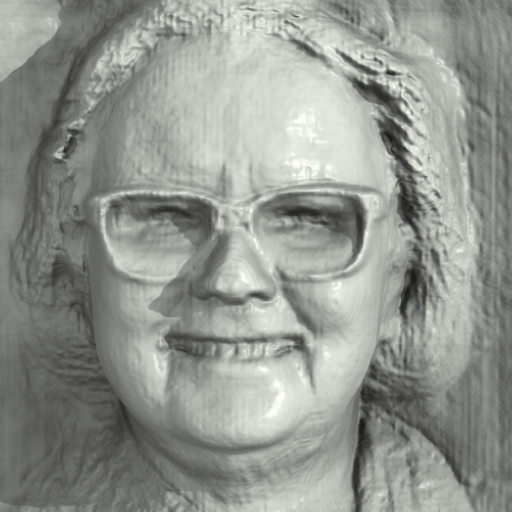}\includegraphics[width=0.2\linewidth]{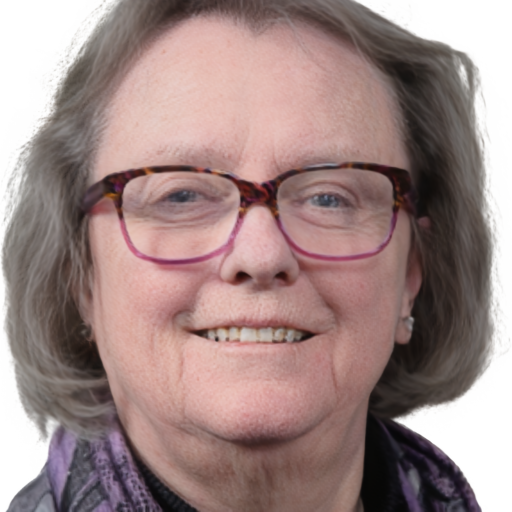}\includegraphics[width=0.2\linewidth]{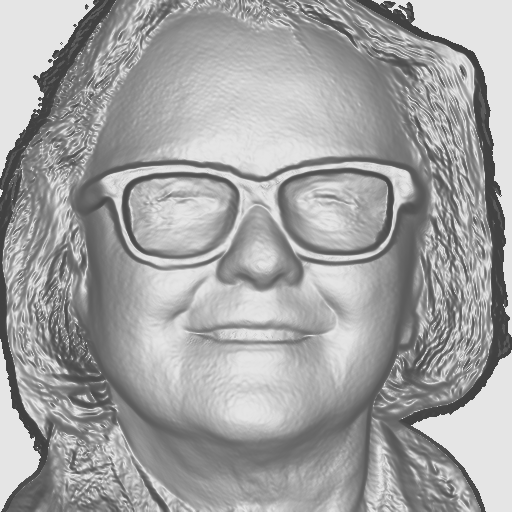} \\
    \caption{Qualitative comparison for images from the FFHQ dataset \cite{karras2019style} showing reconstructed input images, and visualized geometry.}
    \label{fig:qual_ffhq}
\end{figure}

\begin{figure}
    \centering
    \scriptsize
    \begin{tabularx}{0.499\linewidth}{CCC}Input & L3DP~\cite{trevithick2023} & Ours\end{tabularx}\hfill\begin{tabularx}{0.499\linewidth}{CCC}Input & L3DP~\cite{trevithick2023} & Ours\end{tabularx}
    \includegraphics[width=\linewidth]{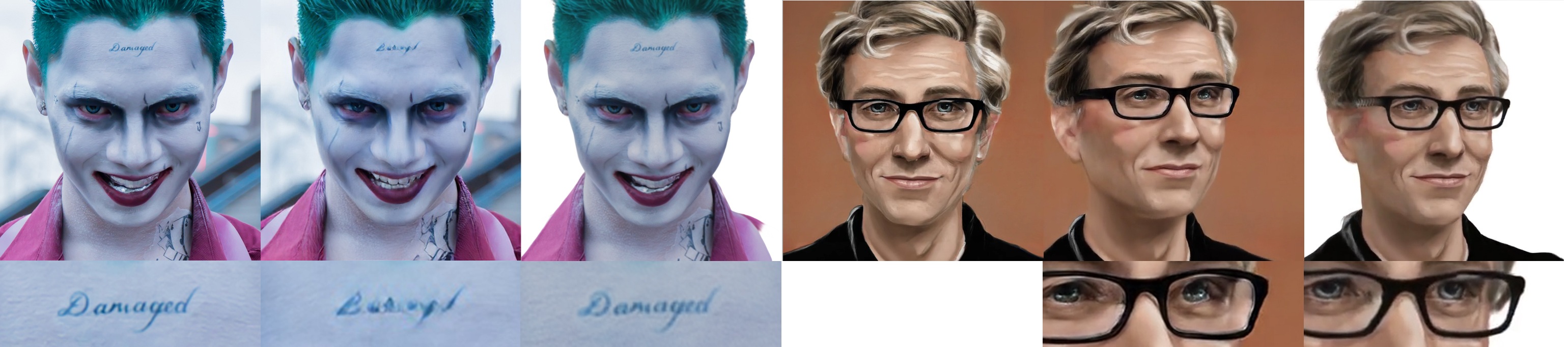}
    \caption{Qualitative comparison for failure case of Live 3D Portrait (L3DP) \cite{trevithick2023}; our method is not constrained by EG3D~\cite{chan2022efficient}, so can authentically represent all details from the input image, and can also create disconnected 3D structures like glasses frames.}
    \label{fig:qual_liv3d_fail}
\end{figure}

We use three datasets for our evaluation.
Firstly, the Cafca dataset \cite{buehler2024cafca} which contains multi-view renders of synthetic humans with variation in identity, expression, clothing and environment.
This dataset provides a diverse range of settings which emulate an in-the-wild context, but with limitations on visual quality and realism.
Secondly, the Ava-256 dataset \cite{martinez2024codec} contains multi-camera captures of people in a studio with controlled clothing and lighting.
This dataset provides less diversity than Cafca, but is not synthetic and so provides far higher realism.
Finally, the FFHQ dataset \cite{karras2019style} includes frontal, in-the-wild photos of a variety of people. 

For Ava-256 and Cafca we follow the protocol of \citet{lyu2024faceliftsingleimage3d} to evaluate novel view rendering (see supplementary material for details). We use PSNR, SSIM, LPIPS and DreamSim metrics to evaluate accuracy of image reconstruction, and ArcFace to assess identity preservation in novel views.
For FFHQ we follow the protocol of \citet{trevithick2023} to evaluate input view reconstruction for a random subset of 500 images.
We use LPIPS, DISTS and SSIM, but omit pose and identity metrics due to insufficient details to enable re-implementation.

\subsection{Results}

\begin{figure*}
    \centering
\scriptsize
\begin{tabularx}{0.495\linewidth}{@{}Q{0.166}@{}Q{0.833}@{}}
     Input frame & 3D reconstruction rendered from -40, -20, 0, +20, and +40 degrees
\end{tabularx}\hfill%
\begin{tabularx}{0.495\linewidth}{@{}Q{0.166}@{}Q{0.833}@{}}
     Input frame & 3D reconstruction rendered from -40, -20, 0, +20, and +40 degrees
\end{tabularx}\\
   \includegraphics[width=0.495\linewidth]{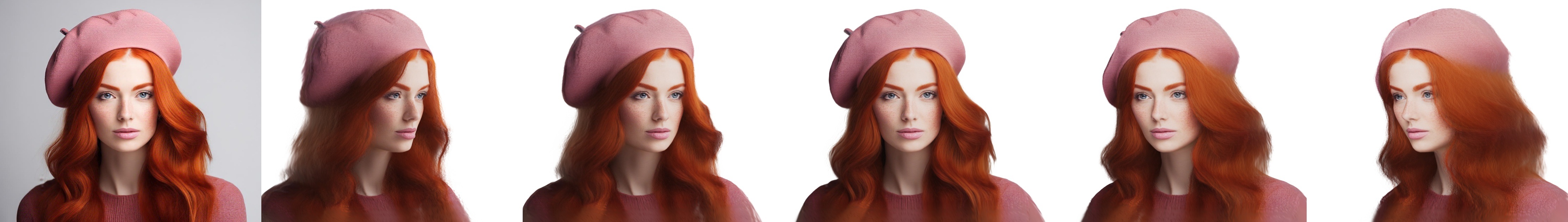}\hfill
   \includegraphics[width=0.495\linewidth]{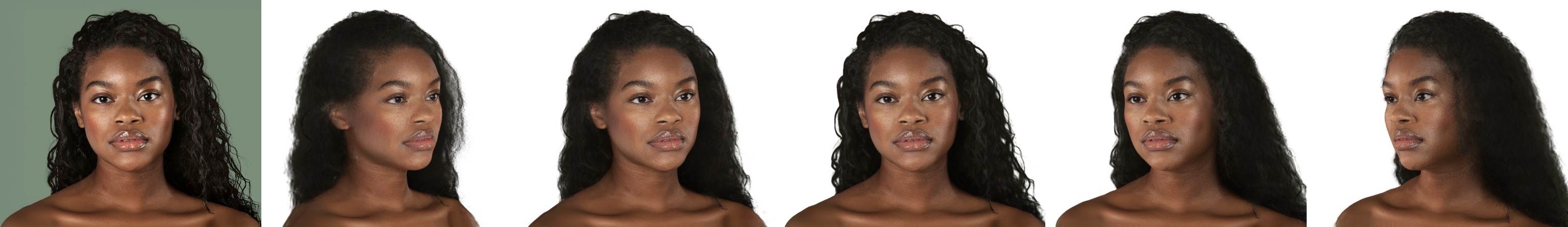}
   \\
   \includegraphics[width=0.495\linewidth]{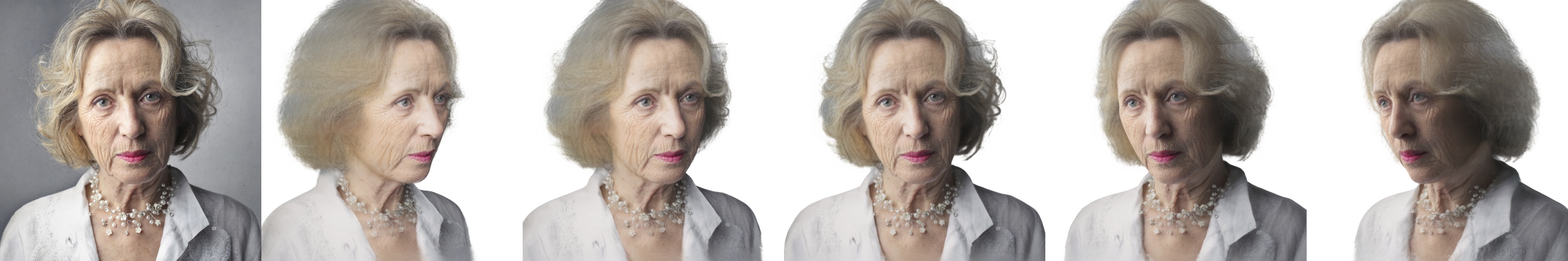}\hfill
   \includegraphics[width=0.495\linewidth]{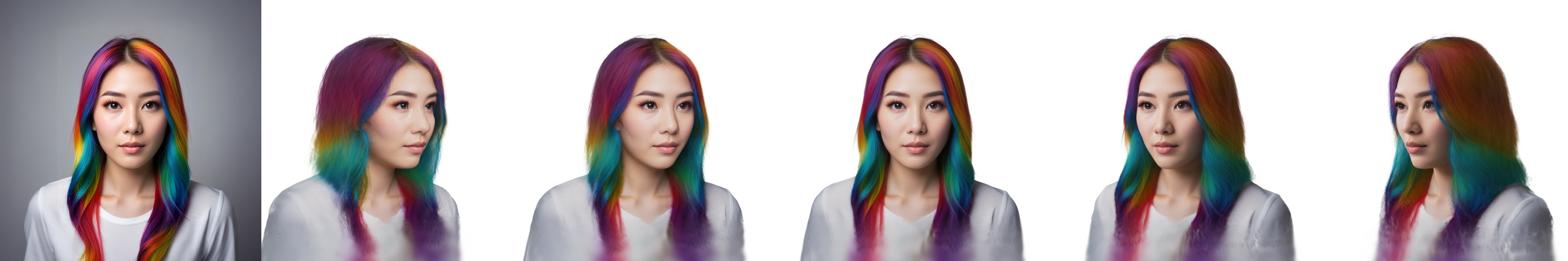}
   % \\
  % \includegraphics[width=0.495\linewidth]{figures/images/qual/0107_supp_wild_v2_left_splitted_7_0.jpg}\hfill
 % \includegraphics[width=0.495\linewidth]{figures/images/qual/0111_supp_wild_v2_right_splitted_4_0.jpg}
   %\\
   %\includegraphics[width=0.495\linewidth]{figures/images/qual/0103_supp_wild_v2_left_splitted_3_0.jpg}\hfill\includegraphics[width=0.495\linewidth]{figures/images/qual/0000_CLO20_Gracie_Megan_001.jpg}
   \\
   \includegraphics[width=0.495\linewidth]{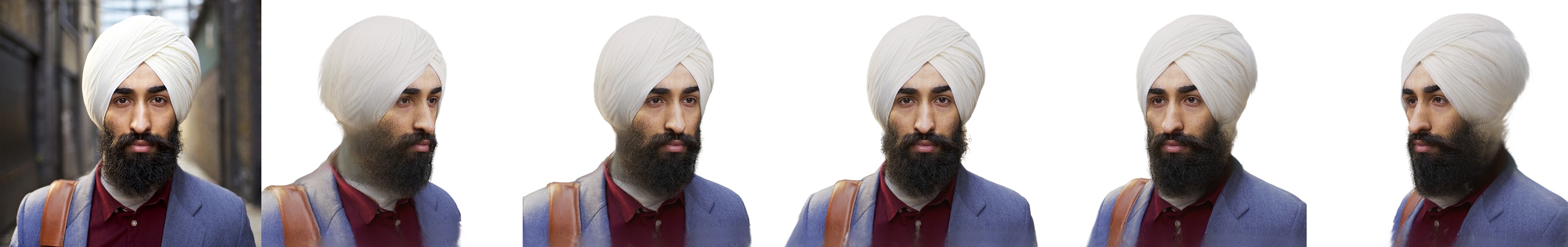}\hfill
   \includegraphics[width=0.495\linewidth]{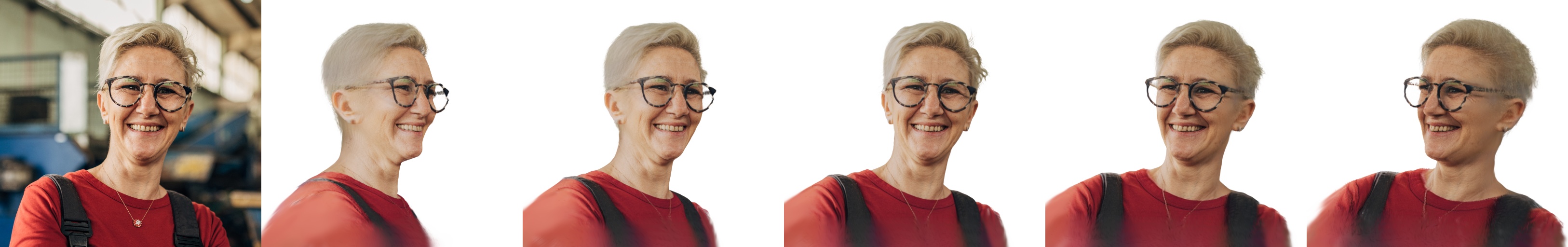}
   \vspace{-0.1cm}
    \caption{Qualitative results of our method. Our approach is able to accurately reconstruct headwear, glasses and other accessories, as well as capturing fine details of hair and skin.}
    \label{fig:qual_res}
\end{figure*}

Results for novel view rendering on the Cafca dataset \cite{buehler2024cafca} are provided in \cref{tab:cafca}. Our method outperforms others both quantitatively and qualitatively, and the results are of high quality.
Compared to FaceLift~\cite{lyu2024faceliftsingleimage3d} our results more accurately preserve the identity of the individual, although the visual quality of our results can degrade more significantly for extreme side views, as shown in \cref{fig:qual_cafca}.
Overall, this produces numerically superior results, and is acceptable for our application where the viewing angle is limited.
Results for novel view rendering on the Ava-256 dataset \cite{martinez2024codec} are provided in \cref{tab:ava_256}.
Results show a similar trend to those for Cafca, with superior identity preservation resulting in better overall metrics. 
See \cref{fig:qual_ava256} for qualitative results.
Qualitative results for in-the-wild images are shown in \cref{fig:teaser,fig:qual_res} and the supplementary material.
Our method performs well across a diverse range of individuals and environments.

Results for input view reconstruction on the FFHQ dataset~\cite{karras2019style} are provided in \cref{tab:ffhq}.
Our method does not perform as well as Live 3D Portrait~\cite{trevithick2023} for DISTS, but is state-of-the-art for LPIPS and SSIM.
Qualitative results shown in \cref{fig:qual_ffhq} demonstrate that the perceived difference in quality is minimal, and that our geometry is smoother and more accurate.
\cref{fig:qual_liv3d_fail} also shows some failure cases of Live 3D Portrait \cite{trevithick2023} which our method is able to address, such as fine details which are not in the EG3D~\cite{chan2022efficient} latent space and disconnected structures such as glasses frames which their approach is not able to represent.
Note that our method is also able to seamlessly fill the `holes' behind occluding objects such as glasses frames.

\begin{table}
  \centering
  \footnotesize
    \caption{Numerical results on Cafca~\cite{buehler2024cafca} subset, following FaceLift protocol \cite{lyu2024faceliftsingleimage3d}. Splatter Image~\cite{szymanowicz24splatter} is retrained on our data.}
\resizebox{\columnwidth}{!}{%
  \begin{tabular}{@{}lccccc@{}}
    \toprule
    Method & PSNR $\uparrow$ & SSIM $\uparrow$ & LPIPS $\downarrow$ & DreamSim $\downarrow$ & ArcFace $\downarrow$ \\
    \midrule
    % PanoHead~\cite{an2023panohead} & 10.72 & 0.7594 & 0.3351 & 0.2048 & 0.2183 \\
    Era3D~\cite{li2024era3d} & 13.69 & 0.7230 & 0.3662 & 0.2892 & 0.2978 \\
   TriPlaneNet~\cite{bhattarai2024triplanenet} & 15.82 & 0.7705 & 0.5303 & 0.1339 & 0.2826 \\
    LGM~\cite{tang2024lgm} & 16.52 & 0.7933 & 0.3060 & 0.1552 & 0.2557 \\
   FaceLift~\cite{lyu2024faceliftsingleimage3d}  & 16.61 & 0.7968 & 0.2694 & 0.1096 & 0.1573 \\
   Splatter Image~\cite{szymanowicz24splatter} & 20.04& 0.8390 & 0.2244 & 0.0276 & 0.1202 \\
   Ours & \textbf{20.39} & \textbf{0.8462} & \textbf{0.1868} & \textbf{0.0286} & \textbf{0.1129} \\
  \bottomrule
  \end{tabular}%
}
  \label{tab:cafca}
\end{table}

\begin{table}
  \centering
  \footnotesize
  \caption{Numerical results on Ava-256~\cite{martinez2024codec} subset, following FaceLift protocol \cite{lyu2024faceliftsingleimage3d}. Splatter Image~\cite{szymanowicz24splatter} is retrained on our data.}
\resizebox{\columnwidth}{!}{%
  \begin{tabular}{@{}lccccc@{}}
    \toprule
    Method & PSNR $\uparrow$ & SSIM $\uparrow$ & LPIPS $\downarrow$ & DreamSim $\downarrow$ & ArcFace $\downarrow$ \\
    \midrule
    Era3D~\cite{li2024era3d} & 14.77 & 0.7963 & 0.2538 & 0.2515 & 0.3721 \\
    TriPlaneNet~\cite{bhattarai2024triplanenet} & 14.32 & 0.6846 & 0.3659 & 0.1692 & 0.2795 \\
    LGM~\cite{tang2024lgm} & 14.05 & 0.8136 & 0.2476 & 0.1496 & 0.3142 \\
    FaceLift~\cite{lyu2024faceliftsingleimage3d} & 16.52 & 0.8271 & 0.2277 & 0.1065 & 0.1871 \\
    Splatter Image~\cite{szymanowicz24splatter} & 18.49 & 0.8258 & 0.2593 &  0.0725 & 0.1458\\
    Ours         & \textbf{19.11} & \textbf{0.8298} & \textbf{0.1908} & \textbf{0.0715} & \textbf{0.1216} \\
  \bottomrule
  \end{tabular}%
}
  \label{tab:ava_256}
\end{table}

\begin{table}
 \centering
 \footnotesize
 \caption{Quantitative evaluation using LPIPS, DISTS, SSIM on 500 FFHQ images~\cite{karras2019style}. Evaluated only using the foreground on $256\times256$ images. * Use a different subset of 500 images; we find the deviation between different subsets to be negligable.}
 \begin{tabular}{@{}l c c c@{}}
 \toprule
 & LPIPS $\!\downarrow$ & DISTS $\!\downarrow$ & SSIM $\!\uparrow$ \\
 \midrule
 ROME*~\cite{Khakhulin2022ROME} & .1158 & .1058 & .8257 \\
 Live 3D Portrait*~\cite{trevithick2023} & .0468 & \textbf{.0407} & .8981 \\
 TriPlaneNet~\cite{bhattarai2024triplanenet} & .2109 & .1580 & .7075 \\
Splatter Image~\cite{szymanowicz24splatter} & .0876 & .0916 & .9165 \\
 Ours & \textbf{.0413} & .0649 & \textbf{.9456} \\
 \bottomrule
 \end{tabular}
 \label{tab:ffhq}
\end{table}

\begin{table}
 \centering
\footnotesize
\caption{Models trained for 100 epochs with and without stability loss $L_j$ with metrics computed with 250 subjects in the Ava256 dataset (more detail in the supplementary material), and our method compared with the Splatter Image method trained for 100 epochs.}
\begin{tabular}{@{}ccccccc@{}}
 \toprule
Method & PSNR $\uparrow$ &SSIM $\uparrow$ &LPIPS $\downarrow$& Jitter $\downarrow$\\
    \midrule
    
Splatter Image  & 18.224 & 0.8651 & 0.2300&0.006353 \\
Ours no $L_j$& 18.949 & 08658&0.1581&0.005020\\
Ours & \bf{19.069}&\bf{0.8682} &\bf{0.1497} &\bf{0.004967}\\
 \bottomrule
  \end{tabular}%
\label{tab:jitter}
\end{table}

Stability is an important attribute for our method. We devise a jitter metric to compare our method to Splatter Image~\cite{szymanowicz24splatter}.
We use images from 250 subjects in the Ava256 dataset with 10 views and 2 frames per subject.
We align the reconstruction and ground truth in the first frame as described by \citet{lyu2024faceliftsingleimage3d} and keep the same 2D alignment transform for the second frame to better measure jitter.
The jitter metric for view $v$ at frame $t$ is defined as $|| (I^{v}_{t}-I^{v}_{t-1}) - (R^{v}_{t}-R^{v}_{t-1})||$. We average this over the views and subjects. % (for only one pair of frames, in this case).
The quantitative results in \cref{tab:jitter} demonstrate that the jitter loss, $L_j$, contributes to improved stability.

\subsection{Limitations}
\begin{figure*}
    \centering
    \scriptsize
    % \begin{tabularx}{\linewidth}{@{}Q{0.166}Q{0.75}}
    %      Input frame & 3D reconstruction rendered from -40, -20, 0, +20, +40 degrees
    % \end{tabularx}
    % \includegraphics[width=\linewidth]{figures/images/failures/facepaint.jpg}\\
    % \includegraphics[width=\linewidth]{figures/images/failures/angle.jpg}\\
    % \includegraphics[width=\linewidth]{figures/images/failures/hands.jpg}\\
    % \includegraphics[width=\linewidth]{figures/images/failures/seg.jpg}

\begin{tabularx}{0.495\linewidth}{@{}Q{0.166}@{}Q{0.833}@{}}
     Input frame & 3D reconstruction rendered from -40, -20, 0, +20, and +40 degrees
\end{tabularx}\hfill%
\begin{tabularx}{0.495\linewidth}{@{}Q{0.166}@{}Q{0.833}@{}}
     Input frame & 3D reconstruction rendered from -40, -20, 0, +20, and +40 degrees
\end{tabularx}\\
    \includegraphics[width=0.485\linewidth]{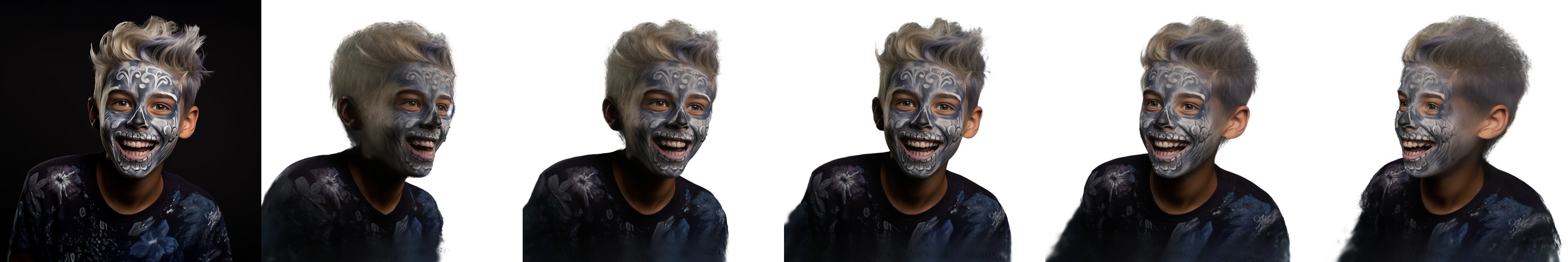}\hfill\includegraphics[width=0.495\linewidth]{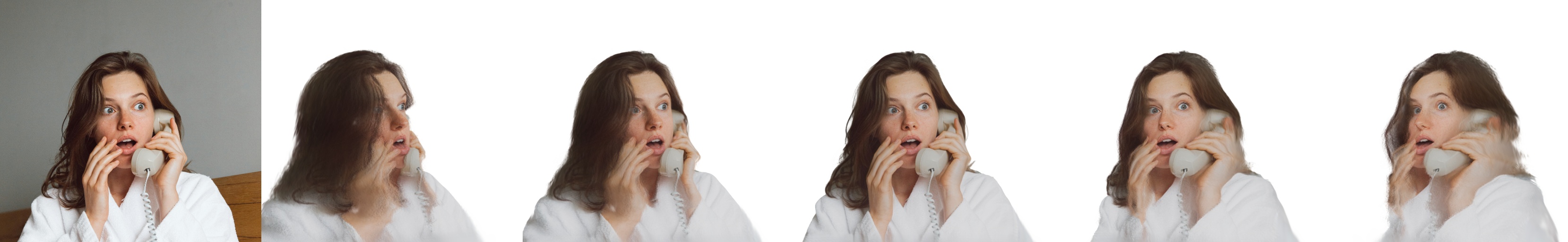}\\
    \includegraphics[width=0.495\linewidth]{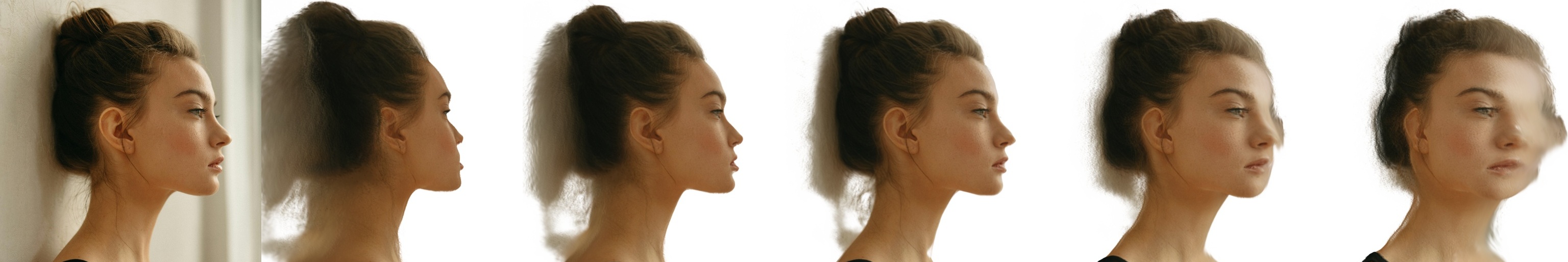}\hfill\includegraphics[width=0.495\linewidth]{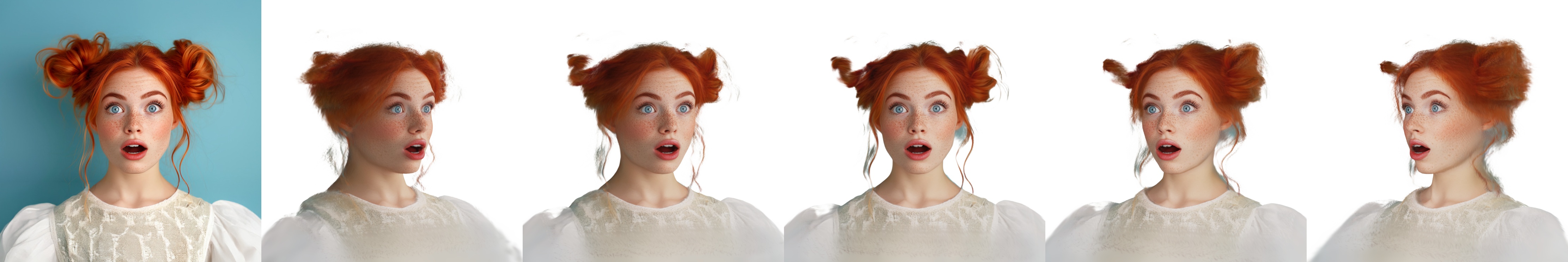}
   \vspace{-0.1cm}
    \caption{Limitations of our method include: unusual texture on the face, extreme input angles, hands/props, segmentation failures. We suspect these are due to lack of representation in the training data; future work will aim to improve the training data to address these issues.}
    \label{fig:limitations}
\end{figure*}

Our method has a number of limitations which largely derive from the training data.
Hands, held objects and face paint are uncommon in the training data and the range of angles of the face and novel view cameras, relative to the input camera, are limited.
As shown in \cref{fig:limitations}, face-paint can result in hallucination of glasses frame structures.
Extreme input angles can cause blurry/visually degraded predictions, although this can be desirable in our scenario to make clear that the prediction in this area is not based on directly observed data.
Reconstruction quality of hands and held objects can be poor when viewed in novel views, and segmentation failures can also occur.
Future work will aim to address these failures through improvements to the training data.

\section{Potential societal impact}

This work aims to bring the experience of remote communication closer to in-person communication, reducing geographic inequality.
By relying only on commodity hardware, we avoid exacerbating
inequality in technology access.
Our method cannot be used to impersonate someone else in an online meeting, due to our requirement for authenticity, supporting trust in online communication and media.
This work also supports authentic representation regardless of factors such as hair colour and type, skin tone, and choice of headwear, aiding diverse self-expression (see \cref{fig:qual_res}).

Like all human-centric computer vision, our 3D reconstruction may have lower accuracy for some demographic groups.
We find that our use of synthetic data is helpful in addressing any lack of fairness we discover, given our precise control over the training data distribution.
Nevertheless, there are aspects of human diversity that are not yet represented by our synthetic datasets (see \cref{fig:limitations}), and there may also be lack of fairness that we have not yet discovered.

\section{Conclusion}

Online communication is growing ever more important, and yet 2D video meetings have challenges that are not solvable within the 2D paradigm.
Recently there has been considerable interest in developing photorealistic avatars that can be controlled and animated from a broad range of signals.
However, we argue that 3D video, with an emphasis on \emph{authentic}, \emph{realistic}, \emph{live}, and \emph{stable} reconstructions, is needed to support online meetings that bring us closer to in-person interactions, while preserving the properties that make videoconferencing so widely adopted today.

In this paper, we presented a method that fulfils the core properties required for authentic 3D video.
There are many opportunities for improvement, and we hope to inspire future work that builds on these results.
With this paper, we hope to highlight an open challenge for the community to demonstrate live 3D reconstruction from monocular inputs that can serve as a true analogue to 2D video, supporting the same diversity of expression that 2D video does today.

\appendix

\section{Evaluation Details}

\label{sec:evaluation_details}
For our evaluations on Cafca and Ava-256 we follow the process of \citet{lyu2024faceliftsingleimage3d}.
For Cafca the renders are 512$\times$512 pixels and include perfect camera parameter annotations.
We use the same subset of the data as \citet{lyu2024faceliftsingleimage3d}, which includes 40 identities with between 9 and 19 views per identity.
For Ava-256 the images are 667$\times$1024 pixels and well-calibrated camera annotations are provided.
Again, we use the same subset of the data as \citet{lyu2024faceliftsingleimage3d}, including 10 identities with 10 views per identity.
The background of the images is removed using the ground-truth masks and replaced by white pixels before we provide the input image to the reconstruction process. 
The ground-truth camera parameters are then used to render several novel views.
\citet{lyu2024faceliftsingleimage3d} align these cameras to their coordinate system while we choose to align our 3D result to the coordinate system of each dataset.
The rendered views and ground-truth images are then aligned using 2D landmark detection to best evaluate the visual quality of the reconstruction, rather than the quality of the alignment.
Note that we only evaluate using the 2D-aligned protocol for Cafca as the unaligned protocol uses all 30 views per subject, these include many views of the rear of the head which we do not target with our method.
For Ava-256 the results are cropped and resized to 512$\times$512 before calculating metrics.
The metrics code is not available so we have re-implemented to the best of our ability with the help of the authors, we do however still expect there to still be discrepancies in alignment/cropping which can have a large impact on the metrics.

\section{Additional Results}

\begin{figure*}[p]
    \centering
\scriptsize
\begin{tabularx}{0.495\linewidth}{@{}Q{0.166}@{}Q{0.833}@{}}
     Input frame & 3D reconstruction rendered from -40, -20, 0, +20, and +40 degrees
\end{tabularx}\hfill%
\begin{tabularx}{0.495\linewidth}{@{}Q{0.166}@{}Q{0.833}@{}}
     Input frame & 3D reconstruction rendered from -40, -20, 0, +20, and +40 degrees
\end{tabularx}
   \includegraphics[width=0.495\linewidth]{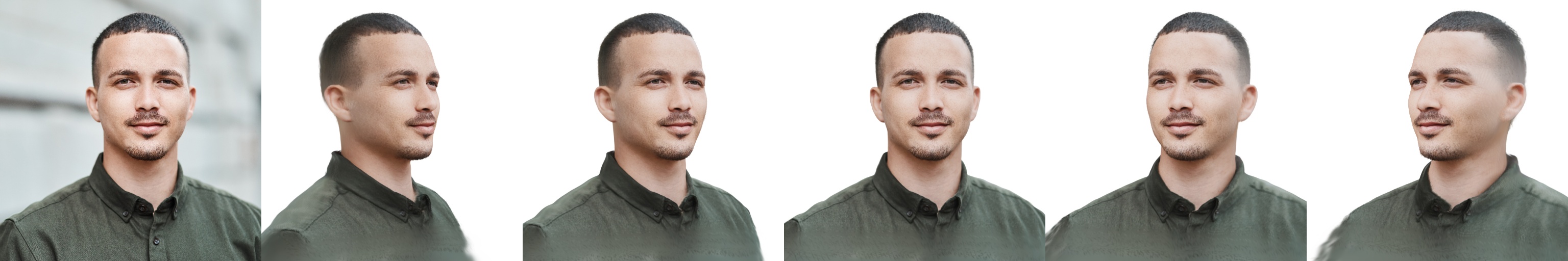}\hfill\includegraphics[width=0.495\linewidth]{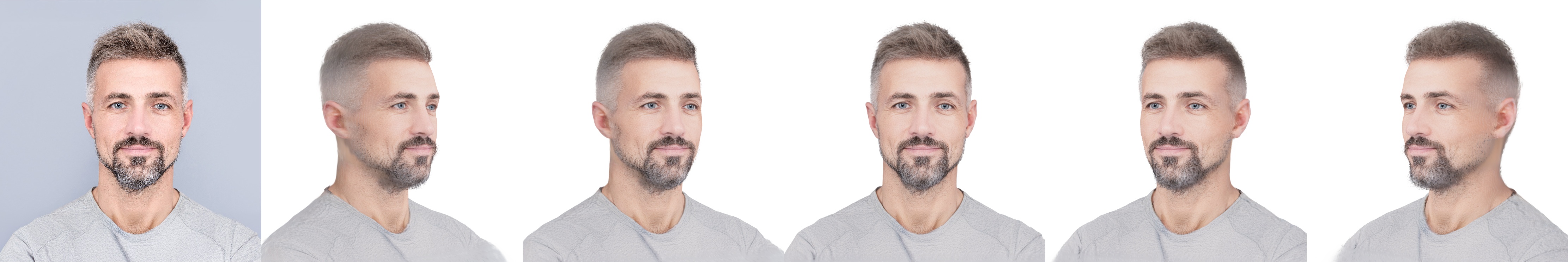}\\
   \includegraphics[width=0.495\linewidth]{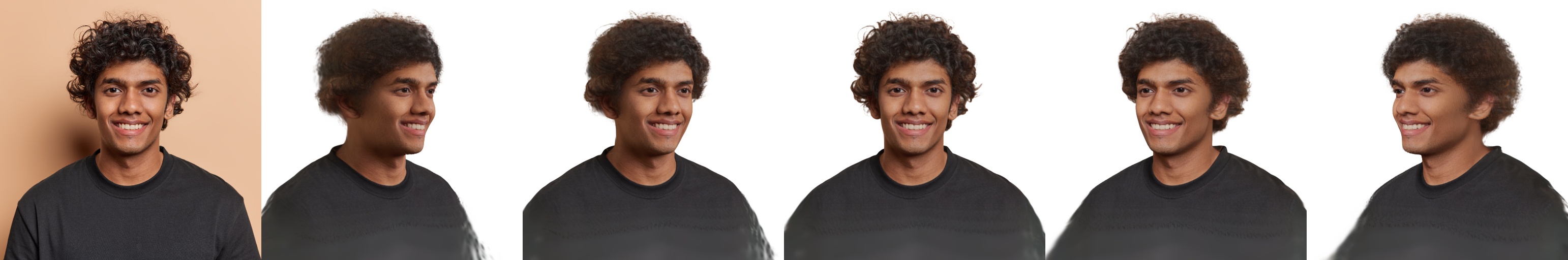}\hfill \includegraphics[width=0.495\linewidth]{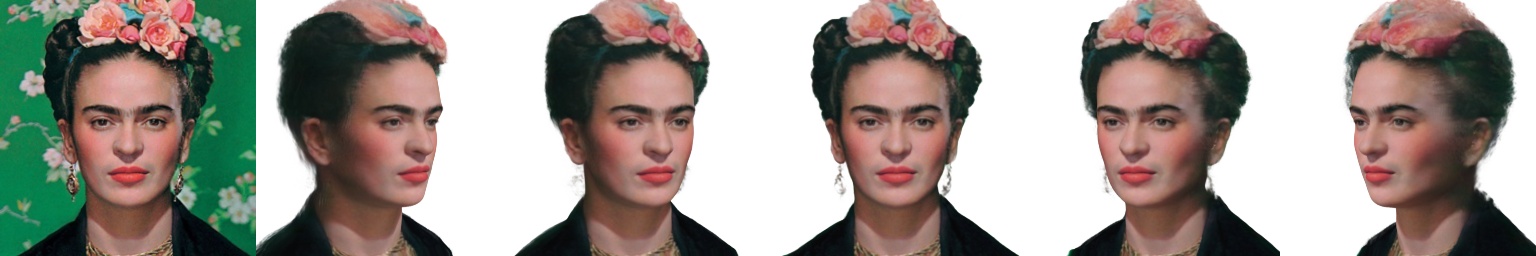}\\
   \includegraphics[width=0.495\linewidth]{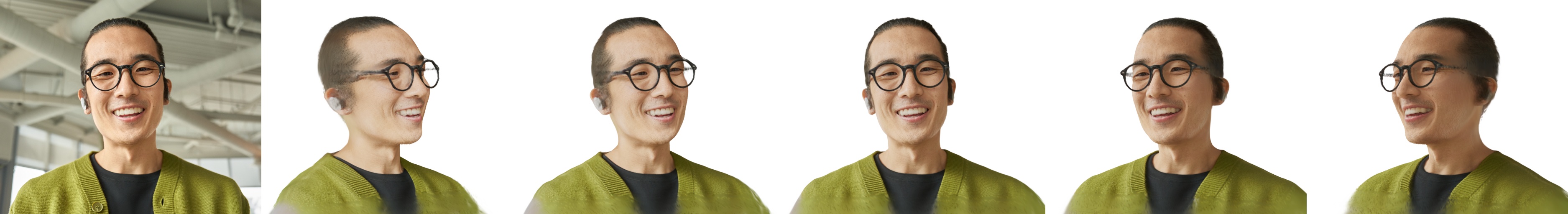}\hfill\includegraphics[width=0.495\linewidth]{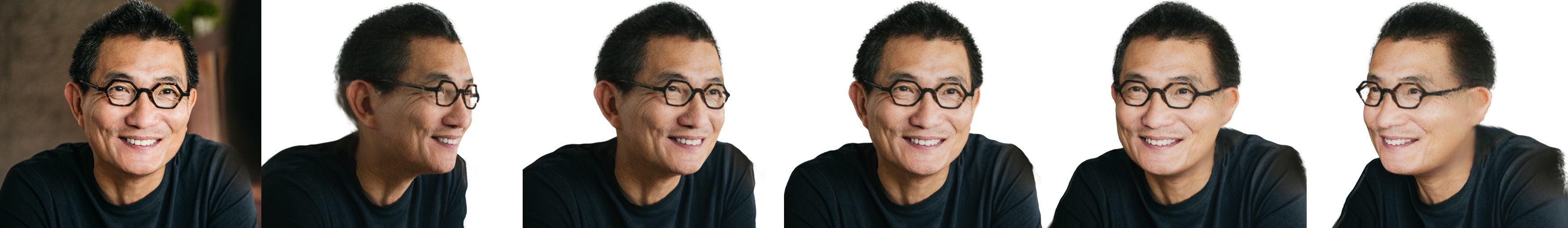}\\
   \includegraphics[width=0.495\linewidth]{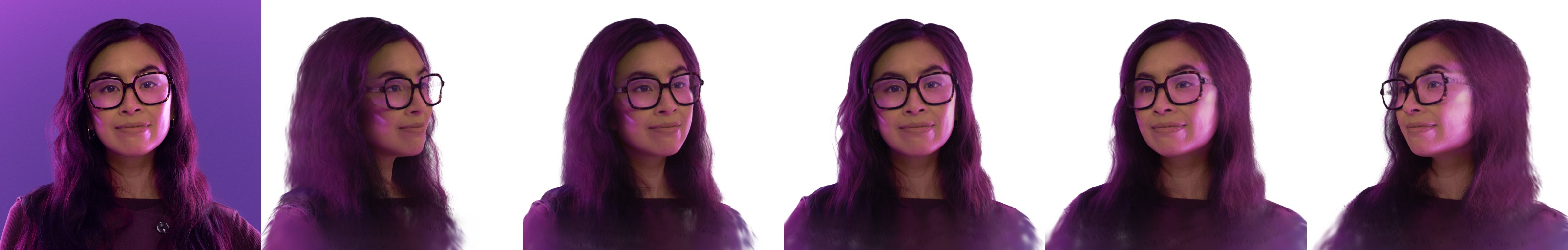}\hfill\includegraphics[width=0.495\linewidth]{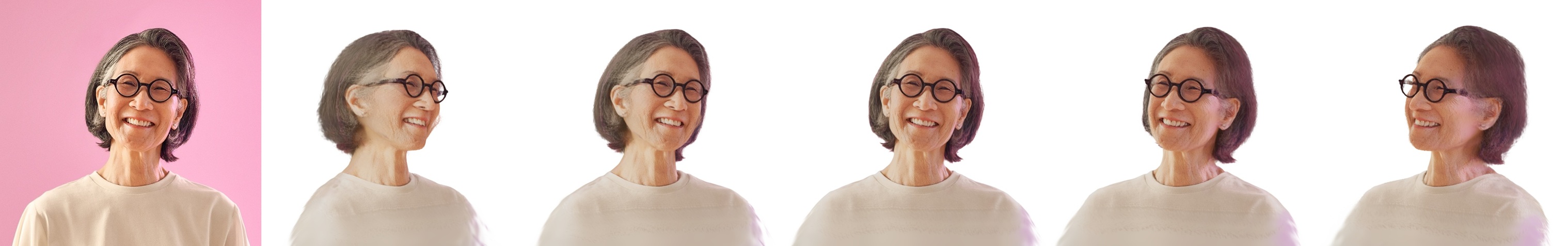}\\
   \includegraphics[width=0.495\linewidth]{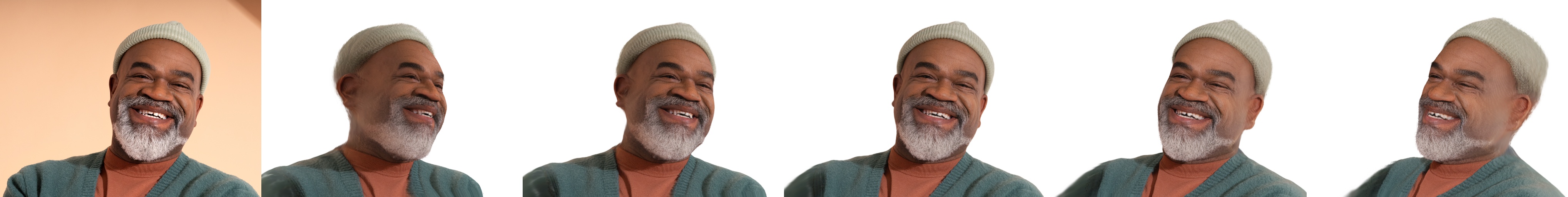}\hfill\includegraphics[width=0.495\linewidth]{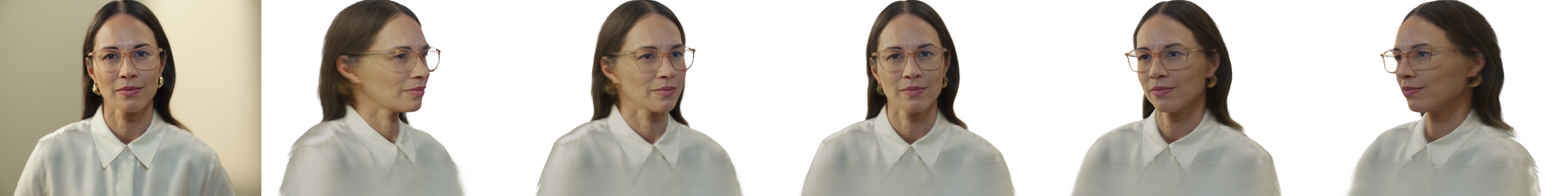}\\
   \includegraphics[width=0.495\linewidth]{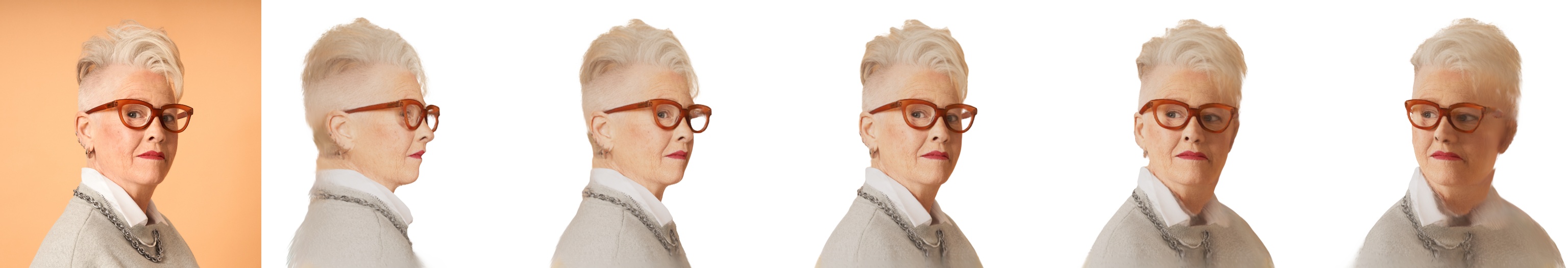}\hfill\includegraphics[width=0.495\linewidth]{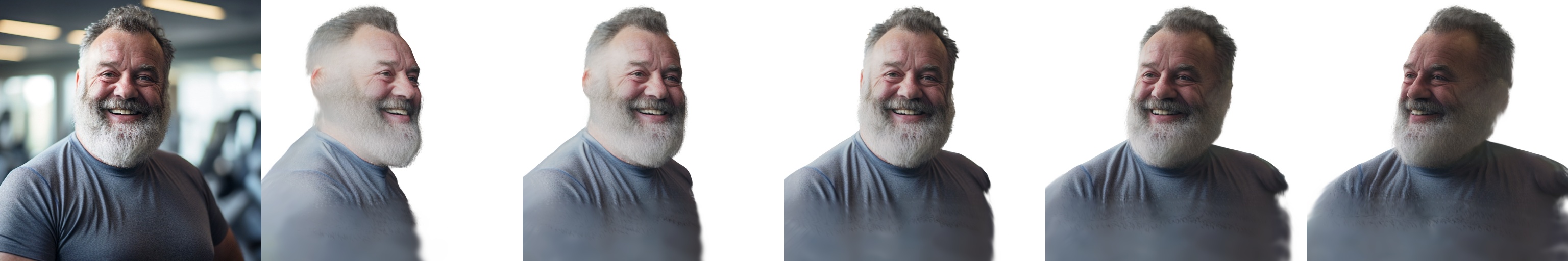}\\
   \includegraphics[width=0.495\linewidth]{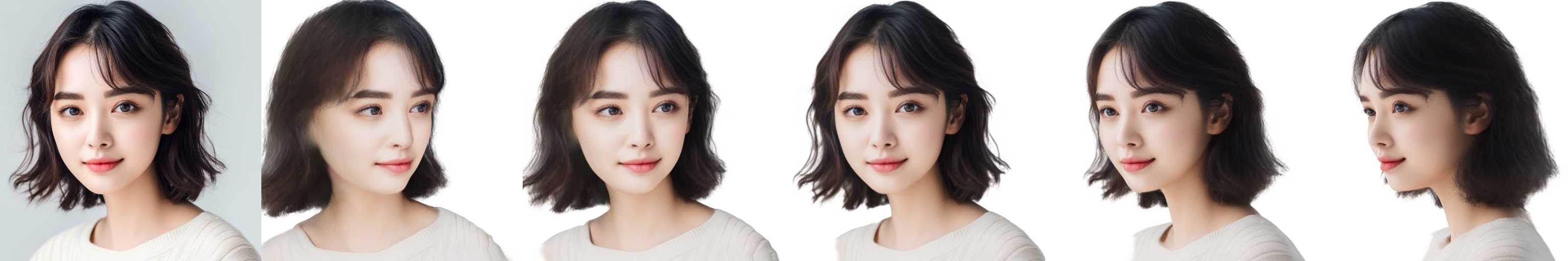}\hfill\includegraphics[width=0.495\linewidth]{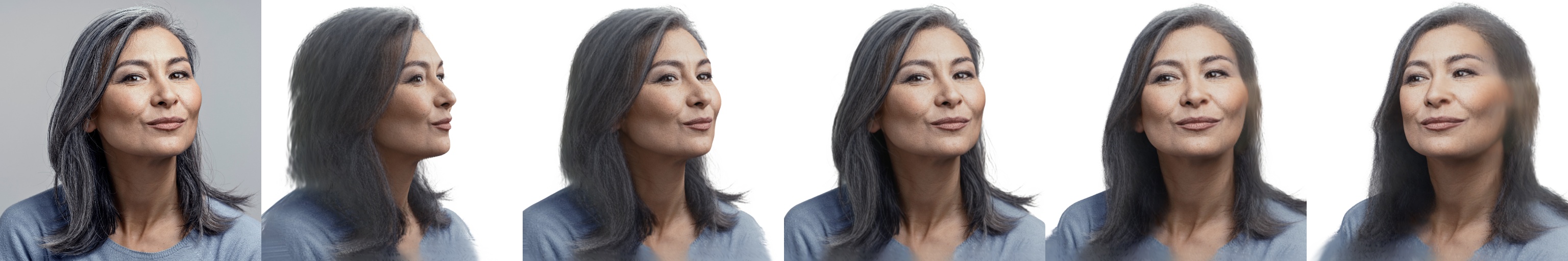}\\
   \includegraphics[width=0.495\linewidth]{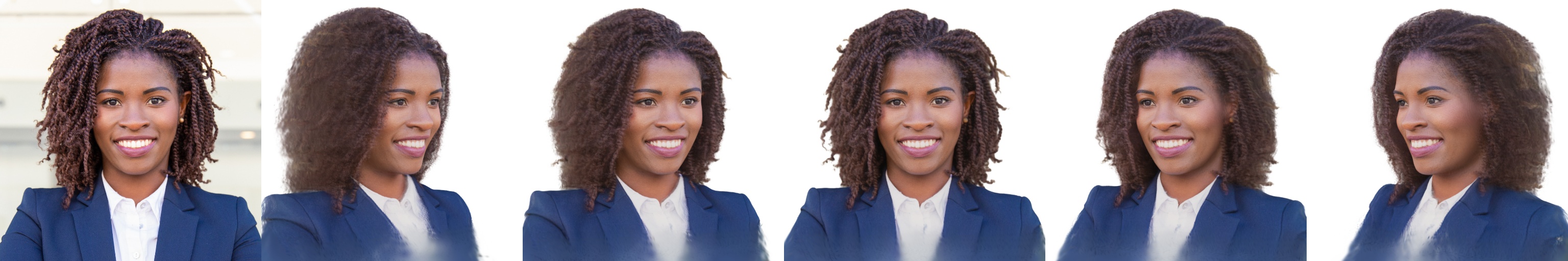}\hfill\includegraphics[width=0.495\linewidth]{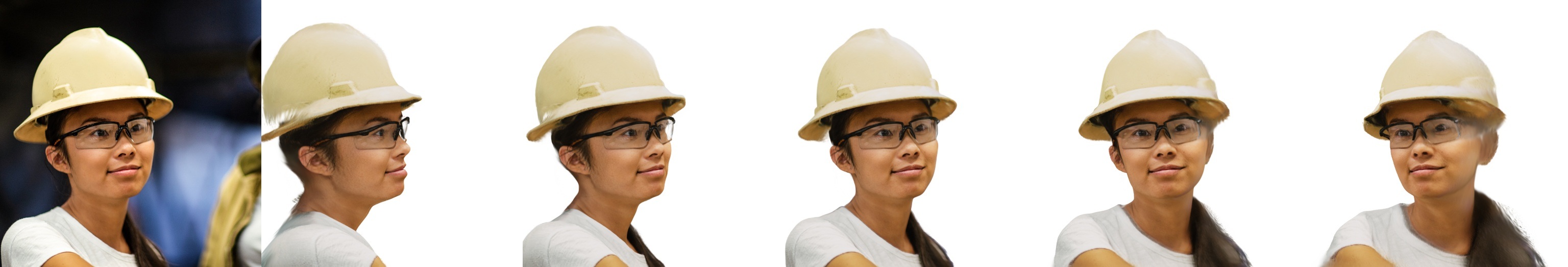}

    \caption{Additional qualitative results of our method.}
    \label{fig:supp_qual_res}
\end{figure*}

\begin{figure*}[p]
    \centering
\scriptsize
\begin{tabularx}{\linewidth}{@{}Q{0.1}@{}Q{0.45}@{}Q{0.45}@{}}
     Input frame & 3D reconstruction rendered from -40, -20, 0, +20, and +40 degrees & Geometry renders in the same views
\end{tabularx}\\
\includegraphics[width=0.545\linewidth]{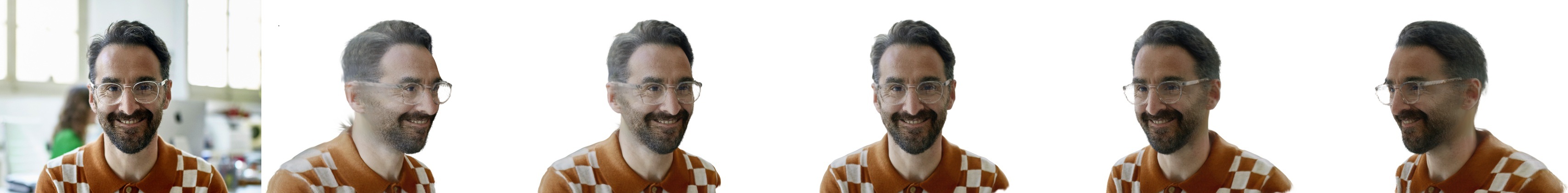}\includegraphics[width=0.454\linewidth]{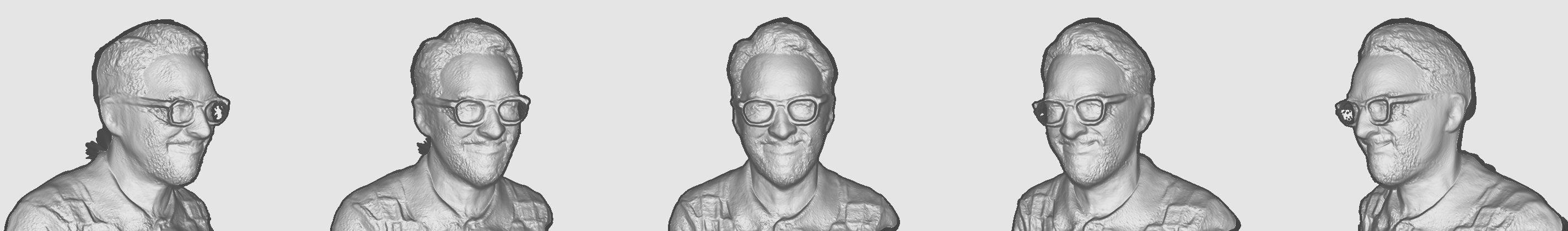}\\
\includegraphics[width=0.545\linewidth]{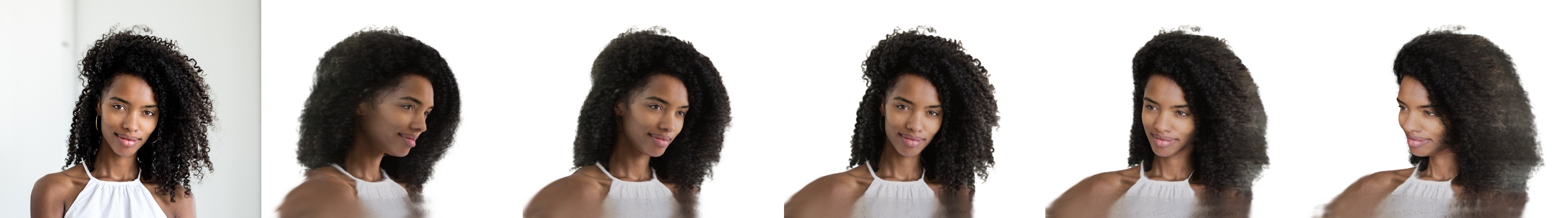}\includegraphics[width=0.454\linewidth]{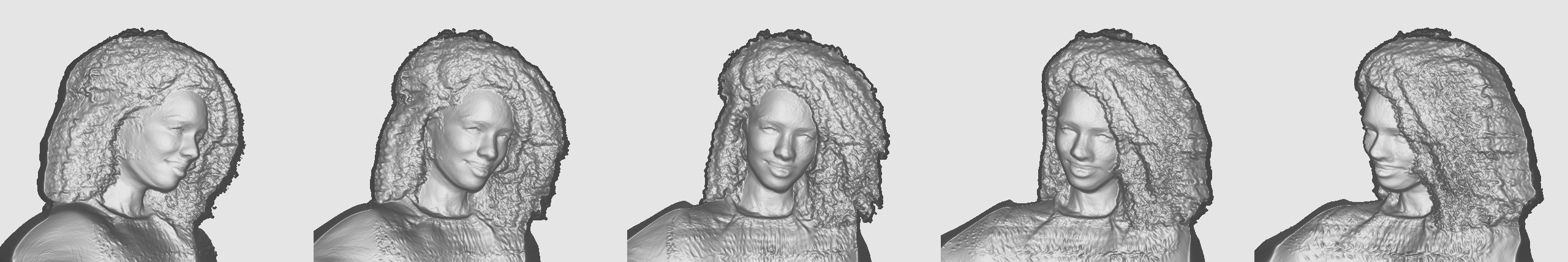}\\
\includegraphics[width=0.545\linewidth]{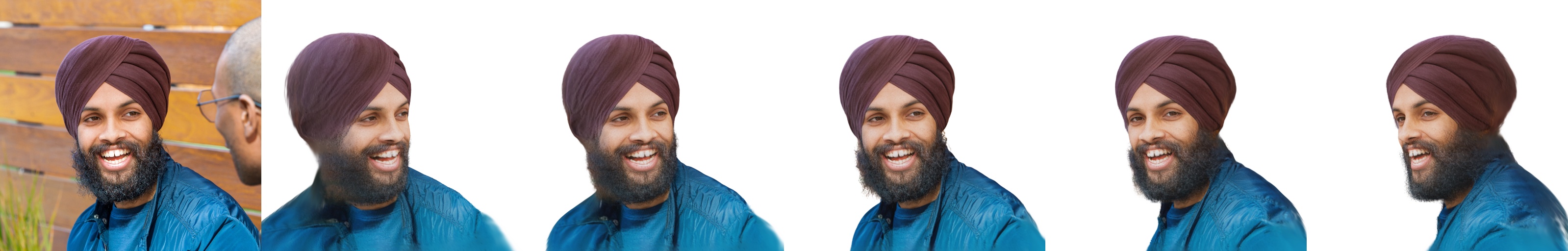}\includegraphics[width=0.454\linewidth]{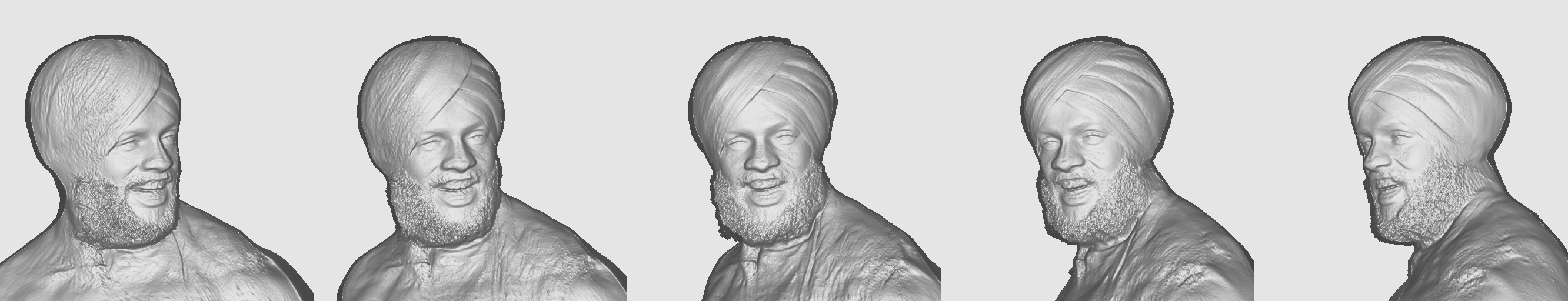}\\
    \caption{Additional qualitative results of our method with geometry renders.}
    \label{fig:supp_geo}
\end{figure*}

\cref{fig:supp_qual_res} provides additional qualitative results of our method.
\cref{fig:supp_geo} shows further results with geometry visualization.
To create these visualizations we render the depth image for Gaussians with slightly larger scale using the Gaussian Splats renderer set up to take the transparencies into account in the alpha-composition. We then calculate the surface normals from this, given the camera intrinsics. 
We then compute an approximate glossy render as a linear combination of the $z$-component of the normal and its square. 
Video results can be found at \url{https://aka.ms/VoluMe}.


\subsection{Impact of multiple Gaussians per pixel}

\begin{figure*}[p]
    \centering
\scriptsize
\resizebox{\linewidth}{!}{%
\begin{tabular}{@{}cc@{}cc@{}c@{}}
     Layer & ~~Input frame & 3D reconstruction rendered from -40, -20, 0, +20, and +40 degrees & ~~Input frame & 3D reconstruction rendered from -40, -20, 0, +20, and +40 degrees\\
     1 & \multicolumn{2}{c}{\includegraphics[width=0.5\linewidth]{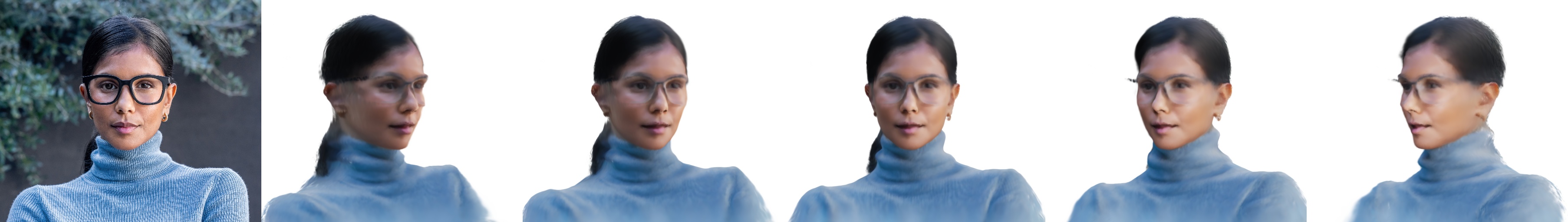}} & \multicolumn{2}{c}{\includegraphics[width=0.5\linewidth]{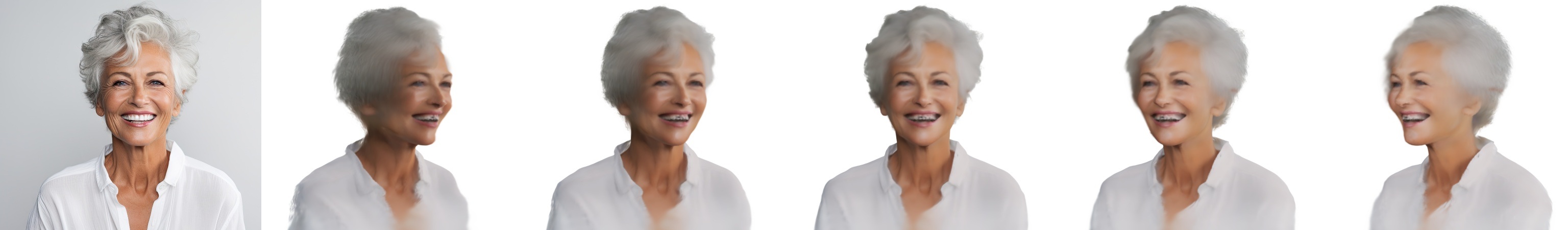}} \\
     2 & \multicolumn{2}{c}{\includegraphics[width=0.5\linewidth]{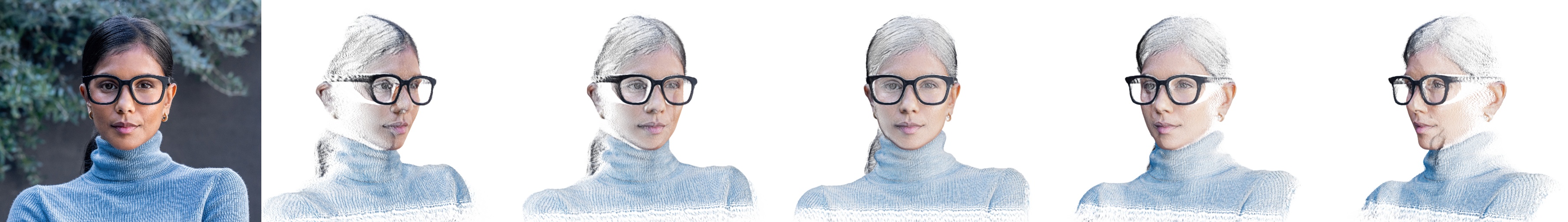}} & \multicolumn{2}{c}{\includegraphics[width=0.5\linewidth]{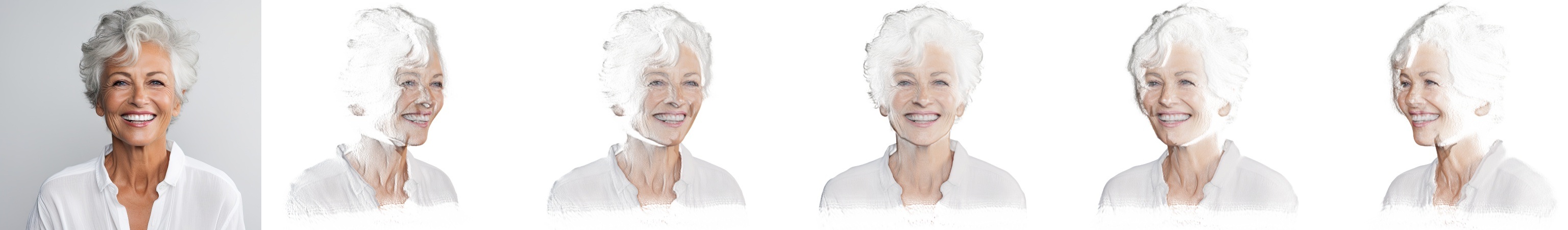}}
\end{tabular}%
}
    \caption{Results showing each `layer' of Gaussians predicted by our method rendered separately. High frequency details and disconnected structures such as glasses are represented by one layer, and the underlying face geometry represented by the other.}
    \label{fig:layers}
\end{figure*}

\cref{fig:layers} shows renders for each of the two Gaussians per pixel that our model predicts in isolation. 
One `layer' of Gaussians is representing the coarse appearance and has clearly learnt a strong prior on the underlying 3D geometry, similar to the results of \citet{szymanowicz24splatter}.
The other layer is behaving more like a monocular depth prediction model, predicting a sheet of small Gaussians which correspond to the first visible surface in the image.
This second layer is therefore able to capture high frequency details that are lacking in previous work, while the overall geometry is still well represented by the first layer.
It is also interesting to note that disconnected structures such as glasses frames are captured well by the second layer, and the first layer can then infill any occluded parts on the face surface behind, based on prior knowledge learnt from the training data. 
The positive impact of using multiple Gaussians per pixel rather than one is also demonstrated quantitatively in \cref{tab:gauss_per_pix}.
Using two Gaussians per pixel improves visual quality at some cost to performance. Using more than two degrades visual quality as measured by the different metrics, and is significantly more costly in terms of performance. We believe this degradation could be due to the difficulty for the network to predict more Gaussians per pixel while keeping the number of channels fixed in the last few layers of the U-Net. Increasing the number of channels in the last few layers could help 3 or 4 Gaussians per pixel work better than 2, but would also come with an increased compute cost.

\begin{table}
\centering
\footnotesize
\caption{Impact of number of Gaussians per pixel. Results on the Ava-256 dataset, following the protocol of \citet{lyu2024faceliftsingleimage3d} with a larger subset of the original datasets and 60 training epochs. All results are with facial landmark alignment.}
\resizebox{\linewidth}{!}{%

\begin{tabular}{@{}ccccccc@{}}
    \toprule
     Gaussians & \multirow{2}{*}{PSNR $\uparrow$} & \multirow{2}{*}{SSIM $\uparrow$} & \multirow{2}{*}{LPIPS $\downarrow$} & \multirow{2}{*}{DreamSim $\downarrow$} & \multirow{2}{*}{ArcFace $\downarrow$} & \multirow{2}{*}{Jitter $\downarrow$}\\
     per Pixel \\
    \midrule
    1 & \underline{19.07}& 0.8668 & \underline{0.1556} & 0.05576 & 0.1114 & \underline{0.004818}\\
    2 &{\bf 19.15}& 0.{\bf 8690} & {\bf 0.1518} & {\bf 0.04677} &  {\bf 0.1003} & 0.005086\\
    3 & 18.99& 0.8663 & 0.1588 & \underline{0.05020} & 0.1073 & {\bf 0.004784}\\
    4 & 19.06& \underline{0.8672} & 0.1567 & 0.05220 & \underline{0.1040} & 0.005990\\
  \bottomrule
\end{tabular}%
}
\label{tab:gauss_per_pix}
\end{table}

\section{Ablation Experiments}

To validate the impact of the various elements of our approach we conduct a series of ablation experiments with results given in \cref{tab:ablations}.

We include a baseline using the simplified U-Net backbone of our method and the same training method as \citet{szymanowicz24splatter}, i.e., where we removed the colour sampling step, the optimized channels and jitter loss, used a single Gaussian per pixel and used simple cropping with CLIFF parameters instead of the homography. We kept for this baseline the re-scaling step to allow training with variable camera-to-face distances as we did not succeed training a model without this term. We provide the results of our full method for comparison against this baseline, and our method omitting each modification as listed above and described in the main paper in sequence.
For these ablation experiments, the original Ava256 dataset subsets used in \citet{lyu2024faceliftsingleimage3d} and described in \cref{sec:evaluation_details} appeared to be too small to allow reliable numerical comparisons between the different methods. We used instead a much larger subset of the original Ava256 datasets. We used 250 samples out of 256 subjects that did not have the camera we chose as input camera. (20220815--1656--MOP211, 20230906--0803--HBM931, 20230817--1136--DZS003,  20230918--1040--NTA876, 20210928--0843--PAK800, 20230310--1106--FCT871) 
For each sequence, the input view was chosen as the camera with index 401168 (which was chosen as the input camera by \citet{lyu2024faceliftsingleimage3d}) when it existed, or the camera 401875 (whose centre was closest to the world origin) otherwise. We then chose the 10 views closest to the input camera centre to compute the metrics. In order to compute the jitter metric we used a pair of consecutive frames for testing. 

Due to time constraints we have been able to train the models for only 6 epochs for this comparison and thus the metrics values are not directly comparable to the values in the other tables in the paper where we used models trained with 100 epochs. However, we observed the ranking between the different methods based on the different metrics to be stable after only 3 epochs and thus we can use these with high confidence to draw conclusions on the relative performance of the different methods.

\begin{table*}
\centering
\footnotesize
\caption{Ablation results on the Ava-256 dataset, following the protocol of \citet{lyu2024faceliftsingleimage3d} with a larger subset of the original datasets and 100 training epochs. All results are with facial landmark alignment. Best results are {\bf bold} and second best \underline{underlined}. The method that combines all our features is first on all metrics but the jitter metric. The baseline method is the worst on all metrics but the jitter metric.}
\resizebox{\linewidth}{!}{%
\begin{tabular}{@{}cccccccccccc@{}}
    \toprule
    \multicolumn{5}{c}{Method} & \multirow{3}{*}{PSNR $\uparrow$} & \multirow{3}{*}{SSIM $\uparrow$} & \multirow{3}{*}{LPIPS $\downarrow$} & \multirow{3}{*}{DreamSim $\downarrow$} & \multirow{3}{*}{ArcFace $\downarrow$} & \multirow{3}{*}{Jitter $\downarrow$}\\
    \cmidrule{1-5}
    Two Gaussian & ROI & Jitter & Direct & Optimizable\\
    Per Pixel & Homography & Loss & Sampling & Layers  \\
    \midrule
    $\times$ & $\times$ &$\times$ & $\times$ &$\times$ & 18.698 &0.8599 & 0.1587 & 0.0553 & 0.1094&\underline{0.004959}\\
    $\checkmark$ & $\checkmark$ & $\checkmark$& $\checkmark$& $\checkmark$ &  \bf{19.069}&\bf{0.8682} &\bf{0.1497} &\bf{0.0468}&\bf{0.0991}&0.004967\\
    $\times$  & $\checkmark$  & $\checkmark$ & $\checkmark$& $\checkmark$        & 18.907&0.8676&0.1528 & \underline{0.0498}&0.1012&\bf{0.004856}\\
    $\checkmark$ & $\times$ & $\checkmark$& $\checkmark$& $\checkmark$        & 18.901 &0.8663 &0.1562&0.0574&0.1018&0.005034\\
    $\checkmark$ & $\checkmark$ & $\times$& $\checkmark$& $\checkmark$        & 18.949 & 0.8658&0.1581&0.0499&0.1027 &0.005020\\
    $\checkmark$ & $\checkmark$ & $\checkmark$& $\times$& $\checkmark$        & \underline{19.026} & \underline{0.8677} &\underline{0.1512}&0.0565&0.0993  &0.004925\\
    $\checkmark$ & $\checkmark$ & $\checkmark$& $\checkmark$& $\times$        & 19.005&0.8660 &0.1585&0.0508&0.1061&0.005170\\

  \bottomrule
  \end{tabular}%
}
\label{tab:ablations}
\end{table*}

The main results from this ablation experiments are: (1) our best method that combines all the different technical contributions of the paper performs much better than the baseline approach across all metrics and performs best or second to best on all but one metric (DreamSim). (2) The model trained without the jitter loss has worse performance on the jitter metric than the model trained with it, which illustrates the benefit of this loss specifically to reduce jitter.

{
    \small
    \bibliographystyle{ieeenat_fullname}
    \bibliography{main}
}

\end{document}